 \documentclass[final,3p,times, 11pt]{elsarticle}

\usepackage{amssymb}
\usepackage{amsmath}
 \usepackage{amsthm}

\usepackage{xcolor}

\newtheorem{theorem}{Theorem}[section]

\newtheorem{definition}{Definition}

\newtheorem{remark}{Remark}

\usepackage{tikz-cd}
\usepackage{multirow}
\usepackage{algorithm}
\usepackage{algorithmic}
\usepackage{subcaption}
\usepackage{cleveref}
\crefname{equation}{}{}

\definecolor{azure}{RGB}{0,127,255}

\makeatletter \def\ps@pprintTitle{\let\@oddhead\@empty \let\@evenhead\@empty \let\@oddfoot\@empty \let\@evenfoot\@oddfoot} \makeatother

\begin{document}

\begin{frontmatter}



\title{Conformal mapping based Physics-informed neural networks for designing neutral inclusions}


\author[1]{Daehee Cho}
\author[1]{Hyeonmin Yun}
\author[2]{Jae Yong Lee\corref{cor1}}
\ead{jaeyong@cau.ac.kr}
\author[1]{Mikyoung Lim\corref{cor1}}
\ead{mklim@kaist.ac.kr}

\cortext[cor1]{Corresponding author}

\affiliation[1]{organization={Department of Mathematical Sciences, Korea Advanced Institute of Science and Technology},
            addressline={291 Daehak-ro, Yuseong-gu}, 
            city={Daejeon},
            postcode={34141}, 
            country={South Korea}}
\affiliation[2]{organization={Department of Artificial Intelligence, Chung-Ang University},
            addressline={84, Heukseok-ro, Dongjak-gu}, 
            city={Seoul},
            postcode={06974}, 
            country={South Korea}}

\begin{abstract}
We address the neutral inclusion problem with imperfect boundary conditions, focusing on designing interface functions for inclusions of arbitrary shapes. 
Traditional Physics-Informed Neural Networks (PINNs) struggle with this inverse problem, leading to the development of Conformal Mapping Coordinates Physics-Informed Neural Networks (CoCo-PINNs), which integrate geometric function theory with PINNs.
CoCo-PINNs effectively solve forward-inverse problems by modeling the interface function through neural network training, which yields a neutral inclusion effect.
This approach enhances the performance of PINNs in terms of credibility, consistency, and stability.
\end{abstract}




\end{frontmatter}



\section{Introduction}
Physics-informed neural networks (PINNs)\cite{Raissi:2019:PINN,karniadakis2021physics} are specialized neural networks designed to solve partial differential equations (PDEs). 
Since their introduction, PINNs have been successfully applied to a wide range of PDE-related problems \cite{cuomo2022scientific,hao2023bilevel,wu2024ropinn}.
A significant advantage of using PINNs is their versatile applicability to different types of PDEs and their ability to deal with PDE parameters or initial/boundary constraints while solving forward problems \cite{NEURIPS2023_8493c860,cho2023separable,pmlr-v235-rathore24a,lau2024pinnacle}.
Balancing the multiple loss terms in PINNs is crucial for efficient training. Recent works have proposed adaptive loss-weighting schemes for PINNs \cite{GAO2025216,BISCHOF2025117914}, in connection with loss-scaling approaches for multitask networks such as GradNorm \cite{chen2018gradnorm}.
The conventional approach to solving inverse problems with PINNs involves designing neural networks that converge to the parameters or constraints to be reconstructed, which are typically modeled as constants or functions. We refer to this methodology as {\it classical PINNs}.
Numerous successful outcomes in solving inverse problems using PINNs have been reported.
See, for example, \cite{Chen:2020:PINN, Jagtap:2022:PINN, Haghighat:2021:PIDL,difonzo2024physics,BERARDI2025117628}.
However, as the complexity of the PDE-based inverse problem increases, the neural networks may require additional design to represent the parameters or constraints accurately.
For instance, \cite{Pokkunuru:2023:ITP} utilized a Bayesian approach to design the loss function, \cite{Guo:2022:MCFP} used Monte Carlo approximation to compute the fractional derivatives, \cite{Xu:2023:TLB} adopted a multi-task learning method to weight losses and also presented the forward-inverse problem combined neural networks, and \cite{Yuan:2022:API} proposed the auxiliary-PINNs to solve the forward and inverse problems of integro-differential equations.
This increase in network complexity can significantly escalate computational difficulties and the volume of data necessary for training PINNs. 
Moreover, the direct approach to approximating the reconstructing parameters via neural networks admits an unconstrained function class, which may lead to instability.
This alludes to the fact that the conventional approach is inadequate depending on the problems, due to the intrinsic ill-posedness of inverse problems.

In this paper, we apply the PINNs' framework to address the inverse problem of designing neutral inclusions, a topic that will be elaborated below. The challenge of designing neutral inclusions falls within the scope where traditional PINNs tend to perform inadequately. To overcome this limitation, we propose improvements to the PINNs approach by incorporating mathematical analytical methods. 

\begin{figure}[htbp!]
\centering
\includegraphics[width=0.5\textwidth]{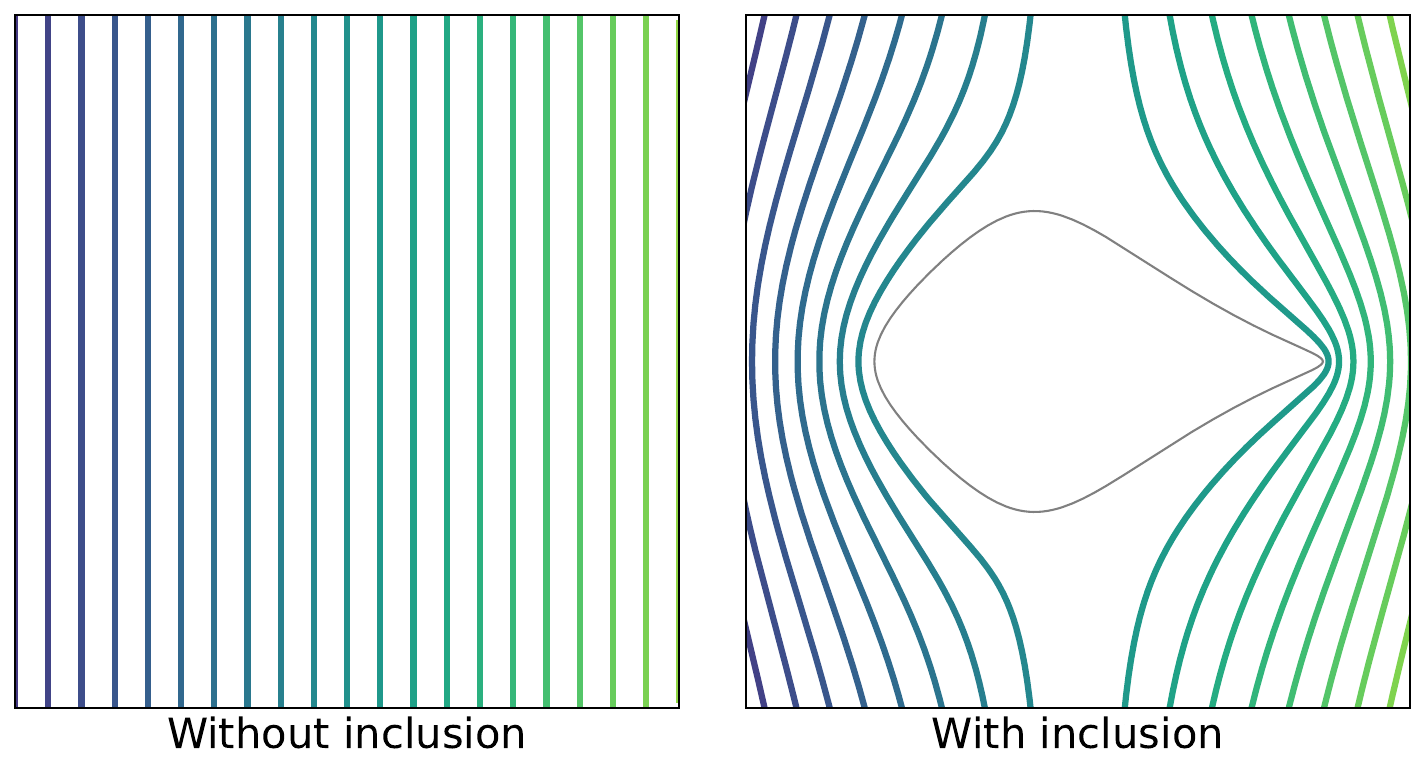}
\caption{
Inclusion-induced perturbation: The inclusion yields a perturbation in the background field, which depends on the interior conductivity, the shape of the inclusion, and the interface conditions.
}
\label{fig:perturbation}
\end{figure}

\begin{figure}[htbp!] 
\centering
\includegraphics[width=\textwidth]{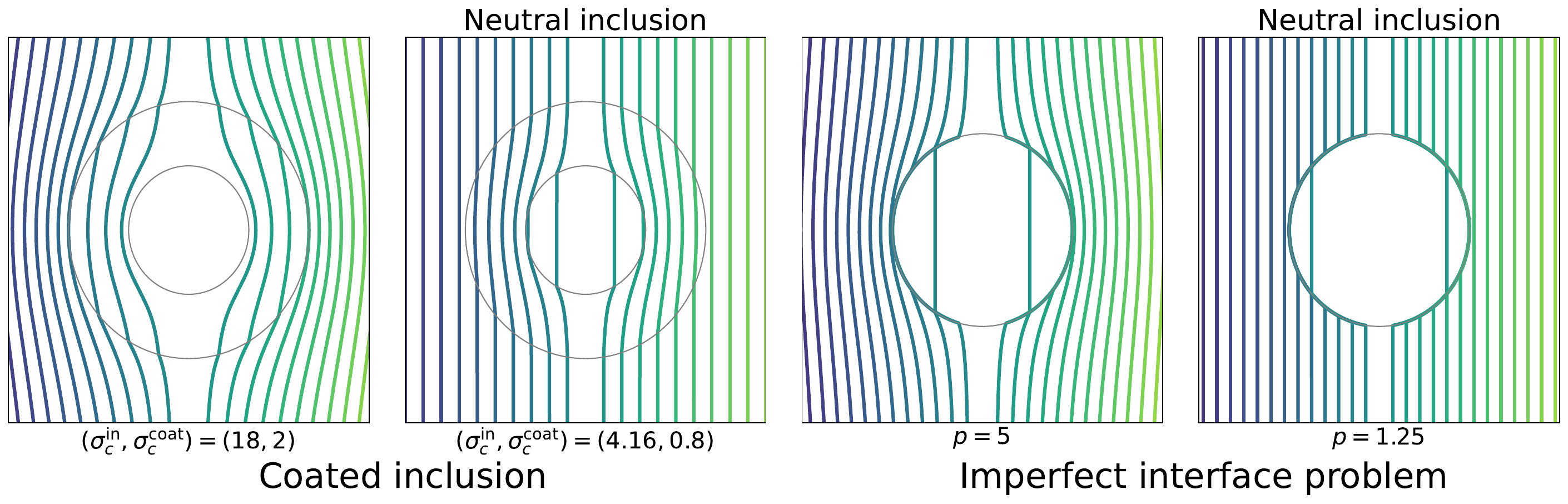}
\vskip -2mm
\caption{
Neutral inclusions with circular shapes: We consider coated circular inclusions with three distinct conductivity regions under perfect bonding conditions, as well as single disk-shaped inclusions with imperfect interface conditions.
}
\label{fig:neural_inclusion}
\end{figure}

Inclusions with different material features from the background medium commonly cause perturbations in applied fields when they are inserted into the medium (see \Cref{fig:perturbation}).
Problems analyzing and manipulating the effects of inclusions have gained significant attention due to their fundamental importance in the modeling of composite materials, particularly in light of rapid advancements and diverse applications of these materials. 
Specific inclusions, referred to as {\it neutral inclusions}, do not disturb linear fields (see \Cref{fig:neural_inclusion}). 
The neutral inclusion problem has a long and established history in the theory of composite materials \cite{Milton:2002:TC}.  
Some of the most well-known examples include coated disks and spheres \cite{Hashin:1962:EMH, Hashin:1962:VAT}, as well as coated ellipses and ellipsoids \cite{Grabovsky:1995:MMEa, Kerker:1975:IB, Sihvola:1997:DPI}. {The primary motivation for studying neutral inclusions is to design reinforced or embedded composite materials in such a way that the stress field remains unchanged from that of the material without inclusions and avoids stress concentration.
Extensive research has been conducted on neutral inclusions and related concepts, such as invisibility cloaking involving wave propagation, in fields including acoustics, elasticity, electromagnetic waves within the microwave range \cite{Alu:2005:ATP, Ammari:2013:ENCm, Landy:2013:FPU, Liu:2017:NNS, Zhou:2006:DEW, Zhou:2007:AWT, Zhou:2008:EWT,Yuste:2018:MIC}.}

Designing neutral inclusions with general shapes presents an inherent challenge. In the context of the conductivity problem, which is the focus of this paper, mathematical theory shows that only coated ellipses and ellipsoids can maintain neutral properties for linear fields in all directions \cite{Kang:2014:CIF, Kang:2016:OBV, Milton:2001:NCI}. In contrast, non-elliptical shapes can remain neutral for just a single linear field direction \cite{Jarczyk:2012:NCI, Milton:2001:NCI}.
To address the difficulty of designing general-shaped neutral inclusions, relaxed versions of the problem have been studied \cite{Choi:2023:GME,Kang:2022:EWN,Lim:2020:IGS}. 
Different from the above examples, where perfectly bonding boundaries are assumed, imperfect interfaces introduce discontinuities in either the flux or potential boundary conditions in PDEs.
\cite{Ru:1998:IDN} found interface parameters for typical inclusion shapes in two-dimensional elasticity for typical inclusion shapes. \cite{Benveniste:1999:NIC} found neutral inclusions under a single linear field.
The interface parameters, which characterize these discontinuities, may be non-constant functions defined along the boundaries of the inclusions so that, theoretically, the degree of freedom of the interface parameters is infinite. Hence, we expect to overcome the inherent challenge of designing neutral inclusions with general shapes by considering neutral inclusions with imperfect interfaces. 
{A powerful technique for dealing with planar inclusion problems of general shapes has been to use conformal mappings and to define orthogonal curvilinear coordinates \cite{Movchan:1997:PSM,Cherkaev:2022:GSE,Ammari:2004:RSI,Jung:2021:SEL},} where the existence of the conformal mapping for a simply connected bounded domain is mathematically guaranteed by the Riemann mapping theory. Using these coordinates, \cite{Kang:2019:CWN,Choi:2024:CIV} constructed weakly neutral inclusions that yield zero coefficients for leading-order terms in PDEs solution expansions (also refer to \cite{Milton:2002:TC,Choi:2023:GME,Lim:2020:IGS} for neutral inclusion problems using the conformal mapping technique).
However, such asymptotic approaches cannot achieve complete neutrality within this framework. Moreover, the requirement for analytic asymptotic expressions poses limitations on the generalizability of this approach.

In this paper, by adopting a deep learning approach, we focus on precise values of the solution rather than asymptotic ones. Unlike the asymptotic approaches in \cite{Kang:2019:CWN,Choi:2024:CIV}, the proposed method does not rely on an analytical expansion formula and incorporates the actual solution values directly into the loss function design.
More precisely, we introduce a novel forward-inverse PINNs framework by combining complex analysis techniques into PINNs, namely Conformal mapping Coordinates Physics-Informed Neural Networks (CoCo-PINNs). 
We define the loss function to include the evaluations of the solutions at sample points exterior to inclusions.
Furthermore, we leverage the conformal mapping to effectively sample collocation points for PDEs involving general-shaped inclusions. 
We found that classical PINNs--which treat the interface parameters as functions approximated by neural networks--perform inadequately when applied to designing imperfect parameters for achieving neutrality. 
Instead, we propose training the Fourier series coefficients of the imperfect parameters, rather than approximating the function. {We test the performance of our proposed method in finding the forward solution by using the analytical mathematical results for the forward solution presented in \cite{Choi:2024:CIV}, where theoretical direct solutions are expressed as products of infinite-dimensional matrices whose entries depend on the expansion coefficients of the interface parameter.

Many PINNs approaches to solving PDEs focus primarily on the forward problem and are typically validated through comparisons with numerical methods such as Finite Element Methods (FEM) and Boundary Element Methods (BEM), and others. 
In contrast, our problem addresses both forward and inverse problems simultaneously, adding complexity, especially in cases involving complex-shaped inclusions, and making it more challenging to achieve accurate forward solutions. Consequently, ``reliability'' becomes a critical factor when applying PINNs to our problem. The proposed CoCo-PINNs provide more accurate forward solutions, along with improved identification of the inverse parameters, compared to classical PINNs. They demonstrate greater consistency in repeated experiments and exhibit improved stability with respect to different conductivity values $\sigma_c$. We conduct experiments to ensure the ``reliability'' of CoCo-PINNs by assessing the credibility, consistency, and stability.

It is noteworthy that by utilizing Fourier series expansions to represent the inverse parameters, CoCo-PINNs offer deeper analytic insights into these parameters, making the solutions not only more accurate but also more explainable. 
Furthermore, our method requires no training data for the neutral inclusion problem due to its unique structure, where constraints at exterior points effectively serve as data. An additional remarkable feature is that our proposed method has proven effective in identifying optimal inverse parameters that are valid for general first-order background fields, including those not previously trained. This impressive result is supported by a rigorous mathematical analysis.

In summary, this paper contains the following contributions:
\begin{itemize}
\item 
We developed a novel approach to PINNs, termed CoCo-PINNs, which demonstrates improved credibility, consistency, and stability compared to classical PINNs in solving forward–inverse problems for conductivity equations with imperfect interface conditions.
\item 
We incorporated the exterior conformal mapping into the PINNs framework to enable training on problems defined over arbitrarily shaped domains.

\item Due to the nature of the neutral inclusion problem and the linearity of the governing equations, training the model with only the two background solutions $H(x) =x_1$ and $H(x) = x_2$ ensures that the neutral effect extends
 to all linear background fields of the form $H(x) = a x_1 + b x_2$, for any $a, b \in \mathbb{R}$ (see \Cref{subsec:untrained_linear} for details).
\end{itemize}

\section{Inverse problem of neutral inclusions with imperfect conditions}
We set $x=(x_1,x_2)$ to be a two-dimensional vector in $\mathbb{R}^2$.
On occasion, we regard $\mathbb{R}^2\cong \mathbb{C}$, whereby $x=x_1+ix_2$ will be used.
We assume that $\Omega\subsetneq\mathbb{C}$ is a nonempty simply connected bounded domain with an analytic boundary (refer \ref{app:ex_conf}).
We assume the interior region $\Omega$ has the constant conductivity $\sigma_c$ while the background medium $\mathbb{R}^2\setminus\overline{\Omega}$ has the constant conductivity $\sigma_m$. We set 
$\sigma = \sigma_c \chi_\Omega + \sigma_m\chi_{\mathbb{R}^2\setminus\overline{\Omega}},$
where $\chi$ is the characteristic function.
We further assume that the boundary of $\partial \Omega$ is not perfectly bonding, resulting in a discontinuity in the potential. 
This discontinuity is described by a nonnegative real-valued function $p(x)$ defined on $\partial \Omega$, referred to as the {\it interface parameter or interface function}.
Specifically, we consider the following potential problem:
\begin{align}\label{eq:main}
\begin{cases}
\nabla\cdot\sigma\nabla u = 0 &\text{in }\mathbb{R}^2,\\[1mm]
p(u|^+ - u|^-) = \sigma_m\frac{\partial u}{\partial \nu}\big|^+ & \text{on }\partial\Omega,\\[1mm]
\sigma_m\frac{\partial u}{\partial \nu}\big|^+ = \sigma_c\frac{\partial u}{\partial \nu}\big|^- & \text{on }\partial\Omega,\\[1mm]
(u-H)(x) = O(|x|^{-1})&\text{as }|x|\to \infty,
\end{cases}
\end{align}
where $H$ is an applied background potential and $\partial u/\partial\nu$ denotes the normal derivative $\partial u /\partial\nu = \langle\nabla u, N\rangle$ with the unit exterior normal vector $N$ to $\partial \Omega$.

Here, we define neutral inclusion to provide a clearer explanation.
\begin{definition}
We define $\Omega$ as a neutral inclusion for the imperfect interface problem \Cref{eq:main} if $(u-H)(x)=0$ for all $x$ in the exterior region $\mathbb{R}^2\setminus\overline{\Omega}$ where $H(x)$ is any arbitrary linear fields, i.e., $H(x) = ax_1+bx_2 $ for any $a, b\in\mathbb{R}$.
\end{definition}
We develop a neural network framework to identify an interface function $p$ that makes the inclusion $\Omega$ a neutral inclusion, given the geometry of $\Omega$ and the conductivities $\sigma_c$ and $\sigma_m$, while also computing the corresponding forward solution $u$.

\subsection{Series solution for the governing equation via conformal mapping}
By the Riemann mapping theorem (see \ref{app:ex_conf}), there exists a unique $\gamma>0$ and conformal mapping $\Psi$ from $D=\{w\in\mathbb{C}:|w|>\gamma\}$ onto $\mathbb{C}\setminus \overline{\Omega}$ such that $\Psi(\infty)=\infty$, $\Psi'(\infty)=1$, and
\begin{equation}\label{def:conformal_sec2}
\Psi(w) = w + a_0 + \frac{a_1}{w}+\frac{a_2}{w^2}+\cdots.
\end{equation}
We set $\rho_0=\ln \gamma$. We use modified polar coordinates $(\rho,\theta)\in[\rho_0,\infty)\times[0,2\pi)$ via $z=\Psi(w)=\Psi(\exp({\rho+i\theta}))$. 
One can numerically compute the conformal mapping coefficients $\gamma$ and $a_n$ for a given domain $\Omega$ \cite{Jung:2021:SEL,Wala:2018:CMD}. 
In the following, we assume $\Psi$ is given. 

\cite{Choi:2024:CIV} derived an explicit analytic solution $u$ to \Cref{eq:main}, denoted by $u_p$, using geometric function theory for complex analytic functions, given an arbitrary analytic domain $\Omega$ and an interface function $p(x)$.
This solution serves as the ground truth for evaluating the credibility of the neural network’s forward prediction $u_{\text{NN}}$, as discussed in \Cref{sec:cred}.

\subsection{Series representation of the interface function}
In this subsection, we derive a series expansion formula for designing the CoCo-PINNs, based on complex analysis.  
We assume that the interface function $p(x)$ is nonnegative, bounded, and continuous on $\partial\Omega$.  
By introducing a parametrization $x(\theta)$ of $\partial\Omega$ for $\theta \in [0, 2\pi)$, the interface function can be expanded into a Fourier series in terms of $\theta$:
\[
p(x(\theta)) =  \sum_{n=-\infty}^{\infty}p_nw^{n},\quad\text{with }|w|=\gamma,
\]
where $p_n$ are complex-valued coefficients.
In particular, we set $x(\theta) = \Psi(w)$, where $w = \gamma \exp({i\theta})$ and $\Psi$ is the conformal mapping defined in \Cref{def:conformal_sec2}.
Since the interface function is real-valued, we have
\begin{align*}
p_{-n} = \overline{p_n}\gamma^{2n}.
\end{align*}
We refer the reader to \cite{Choi:2024:CIV} for further details.
In this case, we have
\begin{equation*}
\begin{aligned}
p(w)&:= p(x(\theta)) 
=p_0 + p_1w + \overline{p_1}\gamma^{2}w^{-1}+
p_2 w^2 + \overline{p_2}\gamma^4 w^{-2}+\cdots,
\end{aligned}
\end{equation*}
for a real-valued constant $p_0$, some complex-valued constants $\{p_k\}_{k=1}^{\infty}$, and $|w|=\gamma$. 
Note that we can similarly express the boundary conditions in \Cref{eq:main} in terms of the variable $w$, enabling us to effectively address these boundary conditions. 

We further assume that $p(w)$ is represented by a finite series, truncated at the $w^n$-term for some $n\in\mathbb{N}$. Specifically, we define
\begin{equation*}
p^{(n)}(w) = \Re\left(\textstyle\sum_{k=0}^n p_k w^k\right).
\end{equation*}
At this stage, the reconstruction parameters are reduced
to $p_0,\cdots,p_n$, and this makes the inverse problem of determining the interface parameter over-determined by using the constraints $(u-H)(x)\approx 0$ for many sample points exterior of $\Omega$.

A fundamental characteristic of inverse problems is that they are inherently ill-posed, and the existence or uniqueness of the inverse solution--the interface function in this paper--is generally not guaranteed. When solving a minimization problem, neural networks may struggle if the problem admits multiple minimizers. In particular, the ability of neural networks to approximate a wide range of functions can lead them to converge to suboptimal solutions, corresponding to local minimizers of the loss function.
As a result, the classical approach in PINNs, which allows for flexible function representation, can be highly sensitive to the initial parameterization. In contrast, the series expansion approach constrains the solution to well-behaved functions, reducing sensitivity to the initial parameterization and ensuring the regularity of the target function. Additionally, since the series approximation method requires fewer parameters, it can be treated as an over-determined problem, helping to mitigate the ill-posedness of inverse problems.

\section{The proposed method: CoCo-PINNs}
This section introduces the CoCo-PINNs, their advantages, and the mathematical reasoning behind why neutral inclusions designed by training remain effective even in untrained background fields.
We begin with the loss design corresponding to the imperfect interface problem \Cref{eq:main}, whose solution exhibits a discontinuity across $\partial \Omega$ due to the imposed boundary conditions.
We denote the solutions inside and outside $\Omega$ as $u^{\text{int}}$ and $u^{\text{ext}}$, respectively, and represent their neural networks' approximations as $u_{\text{NN}}^{\text{int}}$ and $u_{\text{NN}}^{\text{ext}}$.
We named these solutions as trained forward solutions.
We aim to train the interface function, represented by $p^{(n)}$ for the truncated series approximation and $p_{\text{NN}}$ for the fully connected neural network approximation.
The method utilizing $p^{(n)}$ is referred to as CoCo-PINNs, while the approach using $p_{\text{NN}}$ to represent the interface function is called classical PINNs.

\subsection{Model design for the forward-inverse problem}

We utilize three sets of collocation points: $\Omega^{\text{int}},~\Omega^{\text{ext}}$, and $\partial\Omega$, which are finite sets of points corresponding to the interior, exterior, and boundary of $\Omega$, respectively, with a slight abuse of notation for $\partial \Omega$.
We select collocation points based on conformal mapping theory to handle PINNs in arbitrarily shaped domains, and provide a detailed methodology for this selection in \Cref{app:sec:points}.
To address the boundary conditions for $z\in\partial\Omega$, we use $x_z=z+\delta N$ and $y_z=z-\delta N$ for the limit of the boundary from the exterior and interior, respectively, with small $\delta>0$, and unit normal vector $N$.

The loss functions corresponding to the governing equation and the design of a neutral inclusion are defined as follows:
\begin{align}
\label{loss1}
\mathcal{L}_{\text{PDE},u}^{\text{int}}
&= \nabla \cdot \sigma_c\nabla u^{\text{int}}_{\text{NN}}
&&\mbox{for }x\in\Omega^{\text{int}},\\
\label{loss2}
\mathcal{L}_{\text{PDE},u}^{\text{ext}}
&= \nabla \cdot \sigma_m\nabla u^{\text{ext}}_{\text{NN}}
&&\mbox{for }x\in\Omega^{\text{ext}},\\
\label{loss3}
\mathcal{L}_{\text{bd},u}^{(1)} 
&= p^{(n)}(z) \left(u^{\text{ext}}_{\text{NN}}(x_z) - u^{\text{int}}_{\text{NN}}(y_z) \right)
- \sigma_m\frac{\partial u^{\text{ext}}_{\text{NN}} }{\partial \nu}(x_z)
&&\mbox{for }z\in\partial\Omega,\\
\label{loss4}
\mathcal{L}_{\text{bd},u}^{(2)}
&= \sigma_m\frac{\partial u^{\text{ext}}_{\text{NN}} }{\partial \nu}(x_z) -
\sigma_c\frac{\partial u^{\text{int}}_{\text{NN}} }{\partial \nu}(y_z)
&&\mbox{for }z\in\partial\Omega,
\\
\label{loss5}
\mathcal{L}_{\text{Neutral},u} 
&= u^{\text{ext}}_{\text{NN}} - H_u
&&\mbox{for }x\in\Omega^{\text{ext}}.
\end{align}
with $\partial u /\partial\nu = \langle\nabla u, N\rangle$.
In the case where we train using classical PINNs, we replace the interface function $p^{(n)}$ with $p_{\text{NN}}$.

By combining all the loss functions with weight variables $\{w_i\}_{i=1}^5$, we define the total loss with $u$ by
\begin{align}\label{eq:loss:total}
\begin{aligned}
\mathcal{L}_{\text{Total},u} =\ &
\frac{w_1}{|\Omega^{\text{int}}|}\sum_{x\in\Omega^{\text{int}}}
\left(\mathcal{L}_{\text{PDE}}^{\text{int}}\right)^2
+\frac{w_1}{|\Omega^{\text{ext}}|}\sum_{x\in\Omega^{\text{ext}}}
\left(\mathcal{L}_{\text{PDE}}^{\text{ext}}\right)^2\\
\qquad
+&\frac{w_2}{|\partial\Omega|}\sum_{z\in\partial\Omega}
\big(\mathcal{L}_{\text{bd}}^{(1)} \big)^2
+\frac{w_3}{|\partial\Omega|}
\sum_{z\in\partial\Omega}
\big(\mathcal{L}_{\text{bd}}^{(2)} \big)^2
+\frac{w_4}{|\Omega^{\text{ext}}|}\sum_{x\in\Omega^{\text{ext}}}
\left(\mathcal{L}_{\text{Neutral}}\right)^2.
\end{aligned}
\end{align}
Here, $|A|$ denotes the number of elements in the set $A$.
To enforce the non-negativity of the interface function, we introduce an additional loss function $\mathcal{L}_{\text{plus}}=\max\{0,-p\}$.

Define \Cref{loss1,loss2,loss3,loss4,loss5,eq:loss:total} in the same manner, replacing $u$ with $v$ and $H_u$ with $H_v$.
Now, define the total loss by
\begin{align*}
\mathcal{L}_{\text{Total}} = \mathcal{L}_{\text{Total,u}} + \mathcal{L}_{\text{Total},v} + w_5\sum_{x\in\partial\Omega}\mathcal{L}_{\text{plus}}.
\end{align*}

We then consider the following loss with a regularization term:
\begin{align}\label{loss:reg}
&\mathcal{L}_{\text{Reg}} = 
\begin{cases}
\displaystyle\mathcal{L}_{\text{total}} + \epsilon\Big(2\pi \gamma^2|p_0|^2 + 4\pi \gamma^2\textstyle\sum_{k=1}^n (1+k^2)|p_k|^2 \Big),
& p=p^{(n)},\\
\displaystyle\mathcal{L}_{\text{total}}+ \epsilon\|\mathbf{w}_p\|_F^2, & p=p_{\text{NN}},
\end{cases}
\end{align}
where $\mathbf{w}_p$ represents the weights of the neural networks $p_{\text{NN}}$, $\|\cdot\|_F$ is the Frobenius norm, and the $W^{1,2}(\partial \Omega)$-norm is used for $p^{(n)}$, that is,
\begin{align*}
\|p^{(n)}\|_{W^{1,2}(\partial \Omega)}^2
&=\|p^{(n)}\|_{L^2(\partial \Omega)}^2
+\|\nabla p^{(n)}\|_{L^2(\partial \Omega)}^2
=2\pi \gamma^2|p_0|^2 + 4\pi \gamma^2\textstyle\sum_{k=1}^n (1+k^2)|p_k|^2.
\end{align*}
This type of regularization is commonly used to address ill-posed problems.
We used the loss in \Cref{loss:reg} for all experiments.

CoCo-PINNs are designed using complex geometric function theory to address the interface problem. While classical PINNs rely on neural network approximations based on the universal approximation theory, CoCo-PINNs utilize Fourier series expansion, which helps overcome the challenges of ill-posedness in the neutral inclusion inverse problems and ensures that the inverse solution remains smooth. Additionally, this approach allows for the selection of initial coefficients of the interface function using mathematical results.
The results of CoCo-PINNs can be explained by a solid mathematical foundation, as discussed in \Cref{subsec:untrained_linear}. In \Cref{sec:exper}, we examine the advantages of CoCo-PINNs in terms of credibility, consistency, and stability.

\subsection{Neutral inclusion effects for untrained linear fields} \label{subsec:untrained_linear}
In this section, we briefly explain the reason that the CoCo-PINNs can yield the neutral inclusion effect for untrained background fields.
Since the governing equation \Cref{eq:main} is linear with respect to $H$, the following trivially holds by the properties of linear PDEs: 
\begin{theorem}\label{thm1:main}
Consider a domain, denoted by $\Omega$, that is of arbitrary shape and whose boundary is given by an exterior conformal mapping $\Psi(w)$.
If there exists an interface function $p(x)$ that makes $\Omega$ a neutral inclusion for the background field $H(x)=x_1$ and $x_2$, simultaneously, then $\Omega$ is neutral also for all linear fields $H(x)=ax_1+bx_2$ of arbitrary directions $(a,b)\in\mathbb{R}^2$. 
\end{theorem}

By \Cref{thm1:main}, one can expect to find a function $p(x)$ such that $\Omega$ is neutral to all linear fields $H$ by training with only two background fields, assuming such a $p(x)$ exists. Although the existence of this function has not yet been theoretically verified, experiments in this paper with various shapes demonstrate that, for given $\Omega$, there exists a $p(x)$ that produces the {\it neutral inclusion effect}, meaning that the perturbation $u-H$ is negligible for all directions of $H$.

The existence of an interface parameter that induces perfect neutrality (specifically, identifying the class of inclusion shapes that allow for such existence) is a mathematically significant question that remains unresolved.
Since the interface parameters possess infinitely many degrees of freedom, it may be possible to derive analytic relations between the conformal mapping coefficients and the coefficients of the interface parameter that cancel the expansion coefficients of the solution $u$ up to a prescribed order. Such relations would be an important step toward establishing an existence theory for interface parameters that realize neutrality.
Our computational approach—providing numerous numerical examples—could offer valuable insights to aid future analytical investigations into the neutral inclusion problem with imperfect interface.

\begin{remark}
 According to the universal approximation theorem, for a given interface function $p(x)$, the analytic solution $u_p$ on a bounded set and $p$ on $\partial \Omega$ can be approximated by neural networks. Additionally, by the Fourier analysis, $p(x)$ can be approximated by a truncated Fourier series $p^{(n)}$.
 In light of \Cref{thm1:main}, we train using two linear field $H_u = x_1$ and $H_v = x_2$.
\end{remark}

\begin{figure}[htbp!]
\centering
\begin{tikzcd}
&& &  {\color{blue}p^{(n)}: \text{\scriptsize{CoCo-PINNs}}} \arrow{d} & 
\\
& {\color{blue}p_{\text{NN}}: \text{\scriptsize{Classical PINNs}}} \arrow{r}{\text{Univ.}} &
p\arrow{r}{\text{Fourier}}&\{p_k\}_{k=0}^n\arrow{r}{\text{\scriptsize{Analytic solution}}}& u_p\arrow{d}{\text{Cred.}}
\\
&& &  & {\color{blue}u_{\text{NN}}} \arrow{u}
\end{tikzcd}
\caption{
Credibility scheme: The solution $u_{\text{NN}}$ is obtained from either classical PINNs or CoCo-PINNs. The interface function $p_{\text{NN}}$ is identified via classical PINNs, while the truncated series representation $p^{(n)}$ is inferred through the proposed CoCo-PINNs framework.
This scheme enables a direct comparison between the PINN-predicted forward solutions and the analytic solution derived from reconstructed interface parameters, thereby providing a basis for evaluating the credibility of each method.
}
\label{Fig:remarks}
\end{figure}

\Cref{Fig:remarks} demonstrates the operational principles of the neural networks we designed.
We use `Univ.' to represent the universal approximation theorem, `Fourier' for the Fourier series expansion, \cite{Choi:2024:CIV} for the analytical solution derived from the mathematical result, and `Cred.' for credibility.
We note that $u_p$ is the {analytic solution} to \Cref{eq:main} associated with $p$ obtained either from $p^{(n)}$ via CoCo-PINNs or from $p_{\text{NN}}$ via classical PINNs.
The credibility of $u_p$ is evaluated based on the closeness of $u_{\text{NN}}$ to $u_p$, measured by the two norms $\|\cdot\|$ introduced in \Cref{loss:credi,loss:credi2}.
Credibility is a crucial factor in the proposed PINNs' schemes for identifying neutral inclusions with imperfect boundary conditions. 
If the trained forward solution $u_{\text{NN}}$, obtained alongside with $p^{(n)}$ or $p_{\text{NN}}$, is close to $u_p$, we can conclude that the neural networks have successfully identified the interface function, ensuring that $\Omega$ exhibits the neutral inclusion effect. This is because $(u_{\text{NN}}-H)$ has small values in $\Omega^{\text{ext}}$ by the definition of the loss function $\mathcal{L}_{\text{Neutral}}$.
In other words, if the interface function provided by the neural network leads to both a small discrepancy between $u_{\text{NN}}$ and $u_p$, and a small value of the neutral loss $\mathcal{L}_{\text{Neutral}}$, we can consider the interface function to be valid.
However, if $|u_p - u_{\text{NN}}|$ remains large, it becomes unclear whether the discrepancy is due to a failure in reconstructing the interface function (inverse problem) or in predicting the forward solution.
To make this reasoning more precise, observe that
\begin{align*}
\|u_p - H_u\| \leq \|u_p-u_{\text{NN}}^{\text{ext}}\| + \|u_{\text{NN}}^{\text{ext}}-H_u\| = \text{Cred.} + \mathcal{L}_{\text{Neutral},u}.
\end{align*}
Here, the first term $\|u_p - u_{\text{NN}}^{\text{ext}}\|$ quantifies the credibility of the network’s forward prediction (denoted as Cred.), while the second term $\|u_{\text{NN}}^{\text{ext}} - H_u\|$ corresponds to the neutral loss $\mathcal{L}_{\text{Neutral},u}$.
This inequality illustrates that both the credibility and the neutral loss jointly contribute to validating the reconstructed interface function.

The field of AI research is currently facing significant challenges regarding the efficacy and explainability of solutions generated by neural networks. 
Moreover, there is a pressing need to establish ``reliability'' credibility in these solutions.
It is noteworthy that the trained forward solution $u_{\text{NN}}$ deviates from the analytic solution $u_p$ defined with $p_{\text{NN}}$ in several examples, particularly in cases involving complicated-shaped inclusions (see \Cref{sec:cred}).
This discrepancy raises concerns about the ``reliability'' credibility of neural networks.

\section{Experiments}\label{sec:exper}
We present the successful outcomes for designing the neutral inclusions by using the CoCo-PINNs, as well as the experiment results to verify the ``reliability'' in terms of credibility, consistency, and stability.
{\it Credibility} indicates whether the trained forward solution closely approximates the analytic solution, which we assess by comparing the trained forward solution with the analytic solution derived in \cite{Choi:2024:CIV}.
{\it Consistency} focuses on whether the interface functions obtained from the classical PINNs and the CoCo-PINNs converge to the same result for re-experiments under identical environments.
It's worth noting that even if the neural networks succeed in fitting the forward solution and identifying the interface parameter in a specific experiment, this success may only occur occasionally.
Consistency is aimed at determining whether the training outcomes are steady or merely the result of chance, and it can be utilized as an indicator of the steadiness of the training model.
Lastly, {\it stability} refers to the sensitivity of a training model, examining how the model's output changes in response to variations in environments of PDEs.

\begin{figure}[htbp!]
\centering
\includegraphics[width=\linewidth]{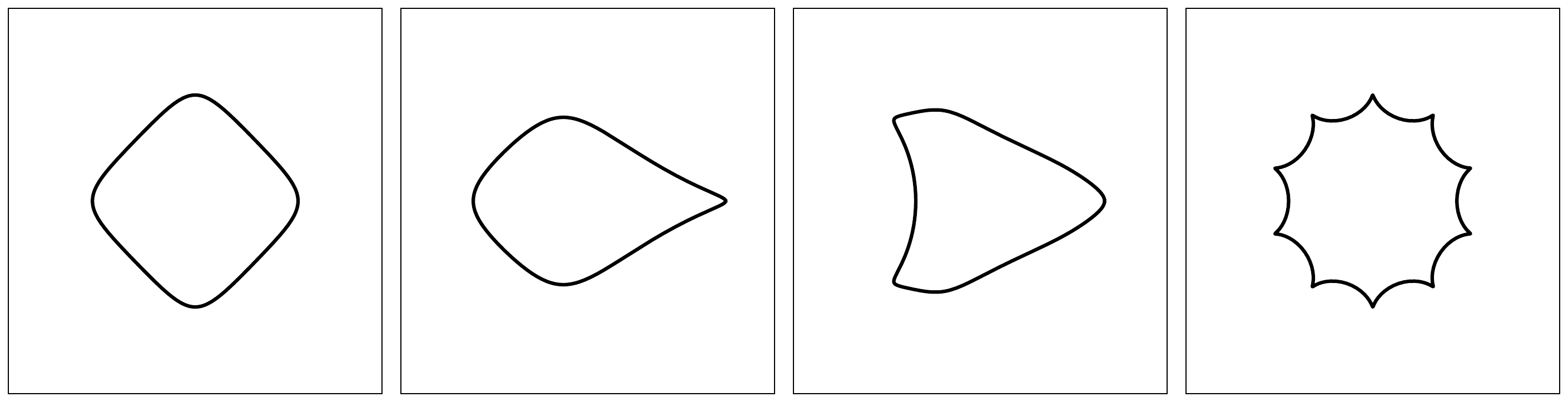}
\caption{
Inclusion shapes: Square, fish, kite, and spike geometries are selected to represent varying degrees of geometric complexity in the assessment.
}
\label{fig:shapes}
\end{figure}

The inclusions with the shapes illustrated in \Cref{fig:shapes} will be used throughout this paper.
Each shaped inclusion is defined by the conformal map given by \Cref{shape:square,shape:fish,shape:kite,shape:custom2} in \Cref{app:shape:info}.
We named the shapes of the inclusions `square', `fish', `kite', and `spike'.

We introduce the quantities to validate the credibility and the neutral inclusion effect as follows:
\begin{align}
\label{loss:credi}
\|u^{\text{ext}}_{\text{NN}}-u_p\|_{\text{Cred}}&= \frac{1}{\left|\Omega^{\text{ext}}\right|}{\sum_{x\in\Omega^{\text{ext}}} |u^{\text{ext}}_{\text{NN}}-u_p|^2 },
\\
\label{loss:credi2}
\|u^{\text{ext}}_{\text{NN}}-u_p\|_\infty &= 
\max_{x\in\Omega^{\text{ext}}}\left\{|u^{\text{ext}}_{\text{NN}}-u_p|\right\},
\\\label{loss:credi3}
\|u_p-H\|_{\text{P-Neutral}}&= \frac{1}{|\Omega^{\text{ext}}|}\sum_{x\in\Omega^{\text{ext}}}
|u_p-H |^2.
\end{align}

\subsection{Experiment setting}
We present detailed descriptions of the experimental setup and corresponding illustrations.
\subsubsection{Collocation points}\label{app:sec:points}
We denote the set of collocation points as 
$$
\Omega^{\text{int}},~\Omega^{\text{ext}},~\partial\Omega,~\partial\Omega^{+},~\partial\Omega^-,
$$
for interior, exterior, boundary, and the limit to the boundary from the interior and exterior components, respectively.
We define it as follows:
$\Omega^{\text{ext}}$ and $\partial \Omega$ are the images of the exterior conformal map $\Psi(w)$ under the uniform grid of its restricted domain, and the boundary of the disk with radius $\gamma=\exp({\rho_0})$ by
\begin{align*}
\Omega^{\text{ext}} 
&= \left\{ \Psi(w): w= \exp({\rho+i\theta}),~(\rho,\theta)\in (\rho_0,L]\times(0,2\pi] \right\},
\\
\partial\Omega
&= \left\{ \Psi(w): w = \exp({\rho_0 + i\theta}),~\theta\in(0,2\pi] \right\},
\end{align*}
for some fixed $L>\gamma$.
The limits to the boundary $\partial \Omega^{\pm}$ from exterior and interior are defined by
\begin{align*}
\partial\Omega^{\pm} = \{z\pm \delta N(z): z\in\partial \Omega\}
\end{align*}
with a small $\delta>0$ and a unit normal vector $N$ to the boundary $\partial \Omega$.
To select the interior points, we recall that  $z \in \partial \Omega$ can be represented as $z = \Psi(\exp({\rho_0 + i\theta}))$.
Now, we fix the angle $\theta$ and divide the radius uniformly. 
In other words, we collect $x \in \Omega$ such that $\arg(x) = \arg(z)$ and $|x| < |z|$, that is,
$$
\Omega^{\text{int}} = 
\left\{ x \in \Omega : |x| < |z|,~ \arg(x) = \arg(z),~ z \in \partial \Omega \right\}.
$$

\subsubsection{Experimental setup and parameters}\label{sec:exp:setting}
In the experimental settings, we set $\sigma_m = 1, \sigma_c = 5 $ and additional values $\sigma_c=3,4,6,7$ are considered.
The conformal radius $\gamma=1$ and $L=5$, which determines the domain of the conformal map. 
Sample points are located within the square $[-5,5]^2$. Collocation points $|\Omega^{\text{ext}}|=21,808, |\Omega^{\text{int}}|=6,000$, and $|\partial\Omega| = 6,000$ were generated once(i.e. fixed throughout the training). 
We employ neural networks with four hidden layers of width 20 and use the Tanh activation function for its smoothness.
Training proceeds for 25,000 iterations with the Adam optimizers under adjusting learning rates \textbf{lr pinn}: all types of neural networks, and \textbf{lr inv}: for interface (inverse) parameters decaying with $\eta$ per 1000 iterations along the learning rate schedulers. 
We fix both learning rates as $10^{-3}$ and $\eta=0.8$.
In the CoCo-PINNs' structure with $p=p^{(20)}$ which takes a real number $p_0$ and complex numbers $\{p_i\}_{i=1}^{20}$ as inverse parameters. To validate the classical PINNs' result, we use the Fourier fitting with an order of $20$. 
The Fourier fitting is explained in \Cref{sec:Fourier:fittings}.
Note that using higher-order terms may introduce singularities, potentially degrading the credibility of the solution. 
For the square and spike-shaped inclusions, the interface function was initialized with a value of 5 to help satisfy the positivity condition more easily.

\textbf{Remark on the environments} \quad Hyperparameter tuning is important for achieving the PINNs' performance. Since balancing them is much more complicated due to the difficulty of the PDEs, we need to choose the appropriate values to enhance PINNs.
Here are the brief guidelines for our settings:
1) \textbf{Collocation Points:} Although the exterior domain is significantly larger than the interior, fitting the background fields in the exterior region is designed to be less complex. The number of collocation points in the exterior domain is set to be four times that of the interior. Since the boundary behavior is crucial for achieving neutral inclusion, the number of boundary points is set equal to that of the interior.
2) \textbf{Fixed Sampling vs. Re-sampling} Re-sampling collocation points can lead to the uncertainty of the interface function. When points are re-sampled, PINNs often struggle to adapt to the new locations, particularly in enforcing boundary condition losses, which are critical for achieving accurate neutral inclusion effects.
Empirical results demonstrate that fixed sampling yields more reliable outcomes; therefore, collocation points remain constant throughout training.
3) \textbf{Adaptive Learning Rates} To control the learning rates adaptively, we use the Adam optimizers so that we can handle the sensitivity of Fourier coefficients much better. Since Adam has faster convergence and robustness, thereby improving stability and accuracy. 

The following algorithm explains the progress of the CoCo-PINNs and classical PINNs.
\begin{algorithm}[htbp!]
    \caption{Generation method of the perturbed field from the interface function}
    \label{alg:computation of interface coefficients}
    {\bf Input: }{Background fields: $H(x, y)$;
    Interface function: $p =p^{(n)}$ or $p_{\text{NN}}$}
\begin{algorithmic}[1]
\IF{$p=p^{(n)}$}
    \STATE{\bf [Initialization]}: Utilizing the mathematical results
    \STATE{Training}
    \STATE $p^{(n)}$ gives us the coefficients $\{p_k\}_{k=1}^n$ directly.
\ELSIF{$p=p_{\text{NN}}$}
        \STATE{\bf [Initialization]}: None
        \STATE $p_{\text{NN}}(w)$ with given sample points $w\in\partial D$.
        \STATE{Training}
        \STATE Use the Fourier fitting to attain the coefficients $\{p_k\}_{k=1}^n$.
    \ENDIF
\STATE { Compute $u_p$ from $\{p_k\}_{k=1}^{n}$ using the analytic solution derived in \cite{Choi:2024:CIV}}.
\end{algorithmic} 
    {\bf{Output:}~}  {$u_p$}
\end{algorithm}

\subsubsection{Conformal maps for various shapes}\label{app:shape:info}
The shapes shown in \Cref{fig:shapes} are defined by the conformal map 
$\Psi(w):\{w\in\mathbb{C}:|w|>1\}\to \mathbb{C}\setminus\overline{\Omega}$ as follows:
\begin{align}
\label{shape:square}\Psi(w) &= w + \frac{1}{10w^3},\\
\label{shape:fish}\Psi(w) &= w + \frac{1}{4w} + \frac{1}{8w^2} + \frac{1}{10w^3},\\
\label{shape:kite}\Psi(w) &= w + \frac{1}{10w}+\frac{1}{4w^2} - \frac{1}{20w^3} + \frac{1}{20w^4} - \frac{1}{25w^5}+\frac{1}{50w^6},\\
\label{shape:custom2}\Psi(w) &= w - \frac{1}{10w^9}.
\end{align}
The \Cref{shape:square,shape:fish,shape:kite,shape:custom2} present `square', `fish', `kite', and `spike', respectively.

\subsubsection{Fourier fittings}\label{sec:Fourier:fittings}
In order to ascertain whether the trained forward solution is true or not, it is necessary to identify the Fourier series that is sufficiently similar to the original interface function and, hence, achieve the real analytic solution.
We utilize the Fourier series approximation for each interface function.
\Cref{fig:Fourier} presents the difference between the interface function and the Fourier series we used.
We denote $p_{\text{F}}$ as the Fourier series corresponding to the $p_{\text{NN}}$.
Given that $p_{\text{NN}}$ is sufficiently close to $p_{\text{F}}$, it is reasonable to utilize the analytic solution obtained by $p_{\text{F}}$ in order to ascertain the credibility of the classical PINNs results.

The relative $L^2$ error of the neural network-designed interface function and its Fourier series formula is given by
$$\frac{\|p_{\text{NN}}-p_{\text{F}}\|_{L^2(\partial\Omega)}}{\|p_{\text{NN}}\|_{L^2(\partial\Omega)}}.$$ 
The relative errors and the Fourier fittings for each shape are given by \Cref{Table:Fourier,fig:Fourier}, respectively.

\begin{table}[t!]\footnotesize
\centering
\caption{The error of Fourier fitting}\label{Table:Fourier}
\renewcommand{\arraystretch}{1.2}
\begin{tabular}{cccc}
\hline
square & fish & kite & spike\\
\hline          
1.122e-03
& 5.192e-04
& 1.535e-03
& 1.347e-03 \\
\hline
\end{tabular}
\end{table}

\begin{figure}[t!]
\centering
\hfil
\begin{subfigure}{0.24\textwidth}
    \centering
    \includegraphics[width=\textwidth]{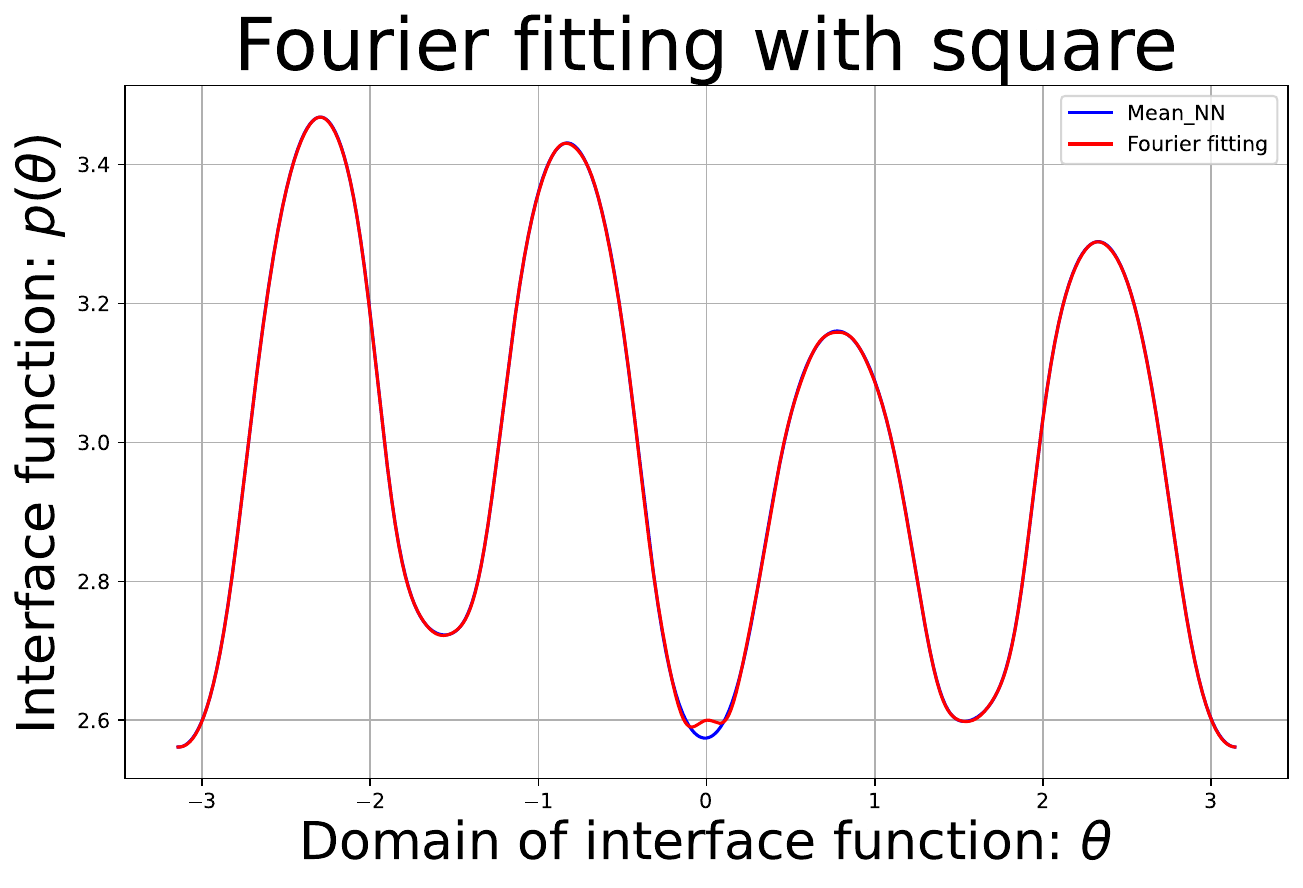}
\end{subfigure}
\hfil
\begin{subfigure}{0.24\textwidth}
    \centering
    \includegraphics[width=\textwidth]{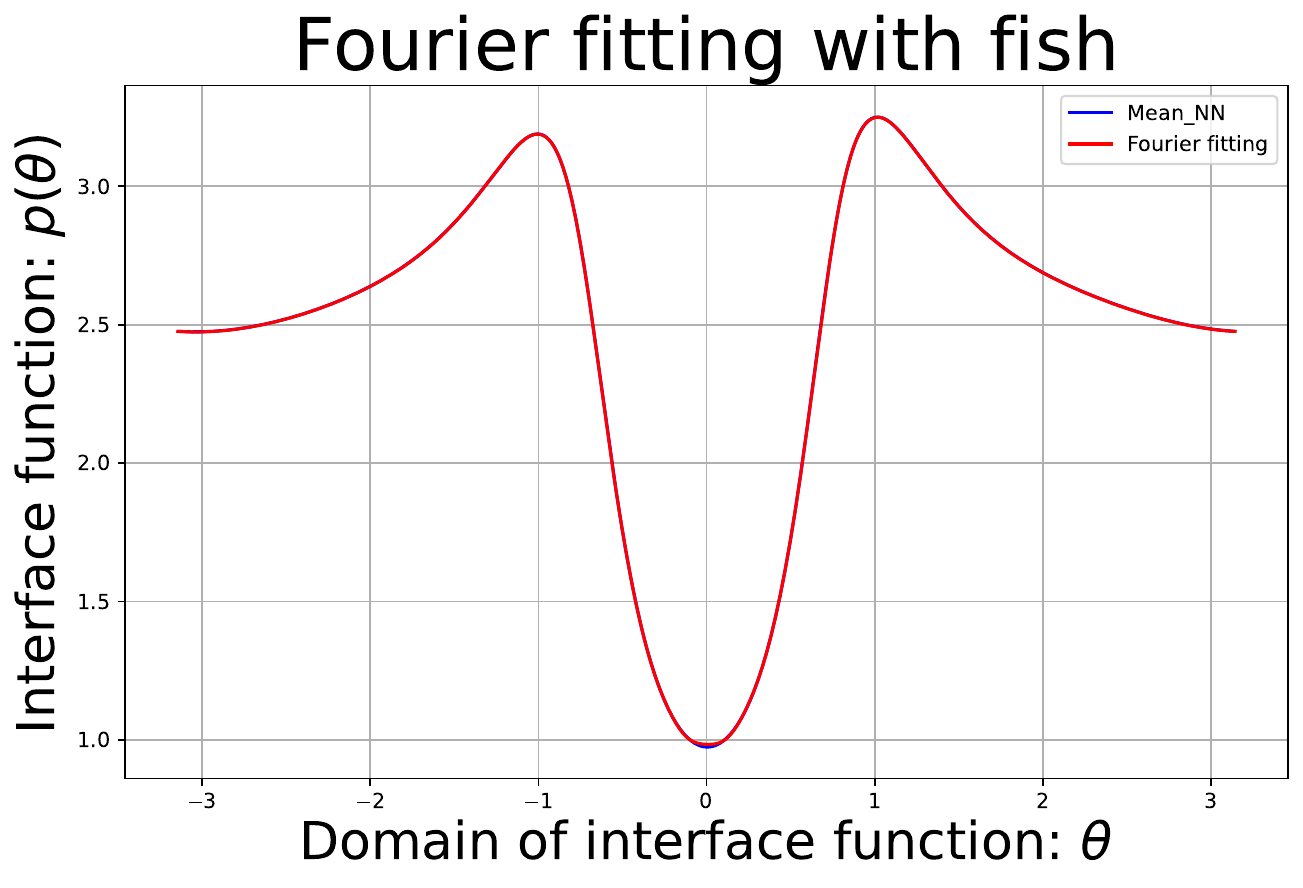}
\end{subfigure}
\hfil
\begin{subfigure}{0.24\textwidth}
    \centering
    \includegraphics[width=\textwidth]{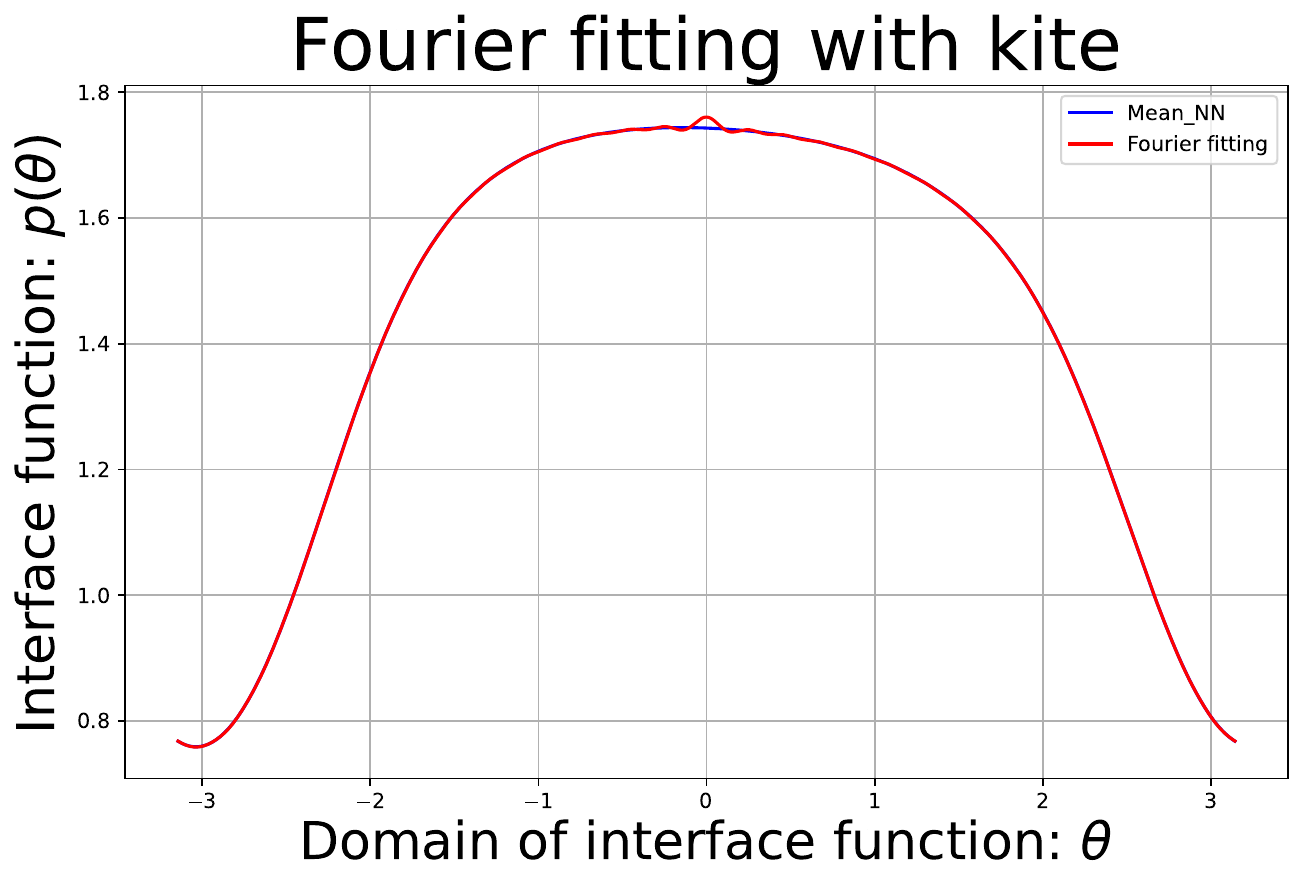}
\end{subfigure}
\hfil
\begin{subfigure}{0.24\textwidth}
    \centering
    \includegraphics[width=\textwidth]{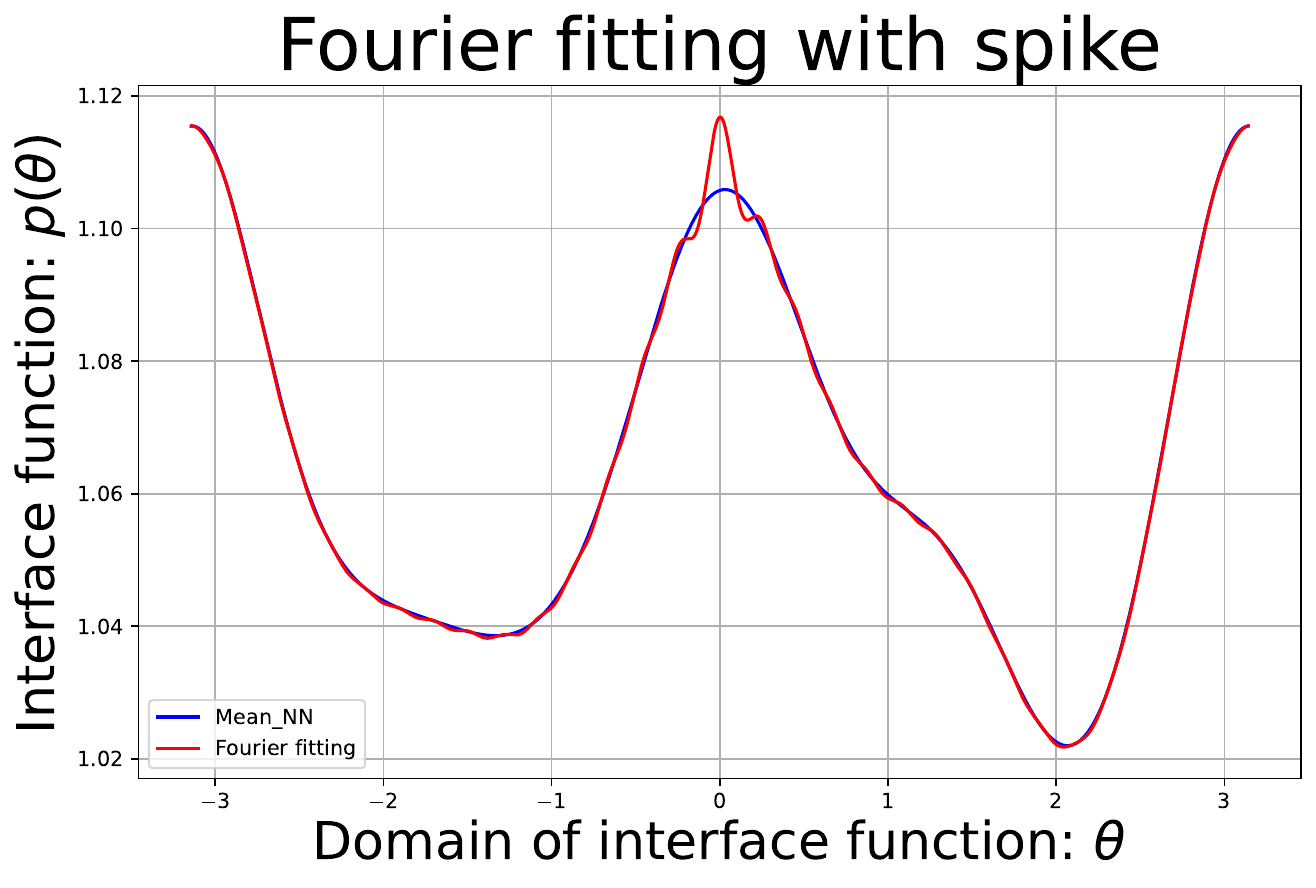}
\end{subfigure}
\hfil
\caption{
Fourier fitting: The interface function obtained by classical PINNs and its Fourier series approximation.
}
\label{fig:Fourier}
\end{figure}

\subsection{Neutral inclusion}\label{sec:neu}
We present experimental results demonstrating the successful achievement of the neutral inclusion effect using CoCo-PINNs.
For training, we use two background solutions, $H_u(x) = x_1$ and $H_v(x) = x_2$. 
We then demonstrate the neutral inclusion effect using three background fields: $H(x) = x_1$, $x_2$, and $ 2x_1-x_2$; see \Cref{thm1:main} for the theoretical justification.

Inclusions generally yield perturbations in the applied background fields.
However, the domain $\Omega$, with the interface function $p(x)$ trained from CoCo-PINNs, achieves the neutral inclusion effects across all three test background fields, as shown by the level curves of the analytic solutions $u_p$ in \Cref{fig:neutral:various}. 

\begin{figure}[t!]
\centering
\includegraphics[width=\textwidth]{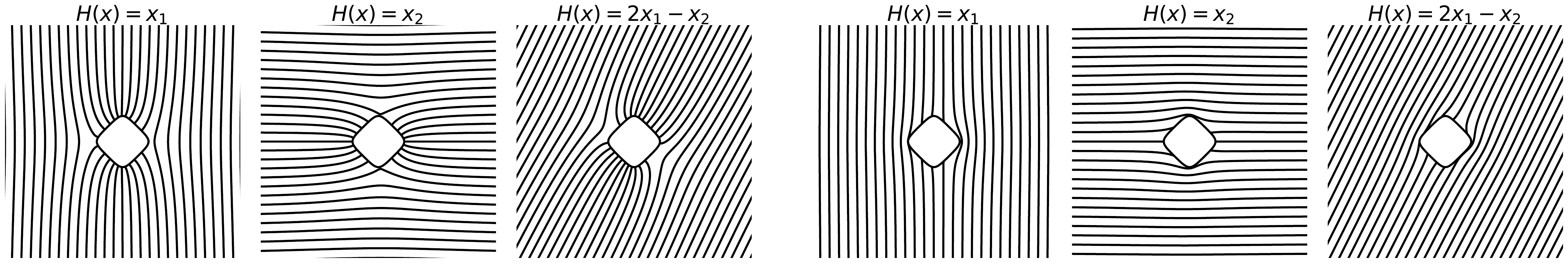}\\
\includegraphics[width=\textwidth]{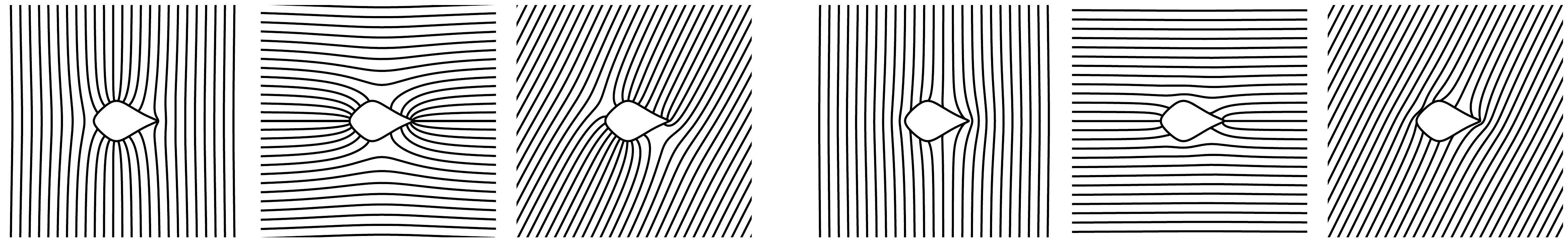}\\
\includegraphics[width=\textwidth]{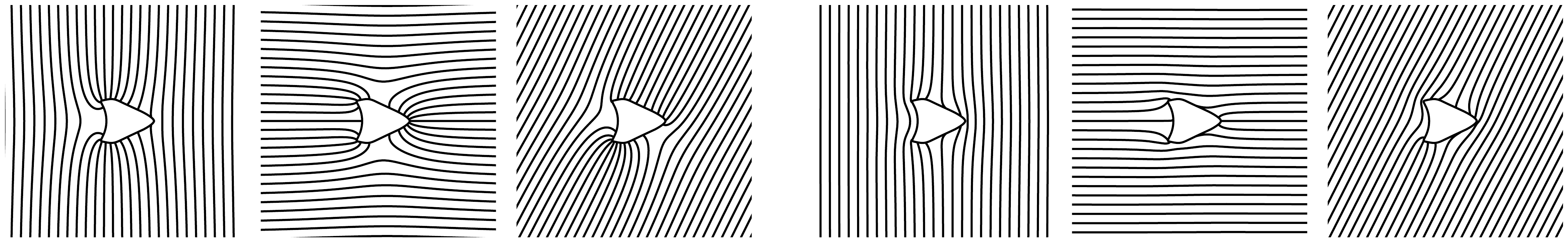}\\
\includegraphics[width=\textwidth]{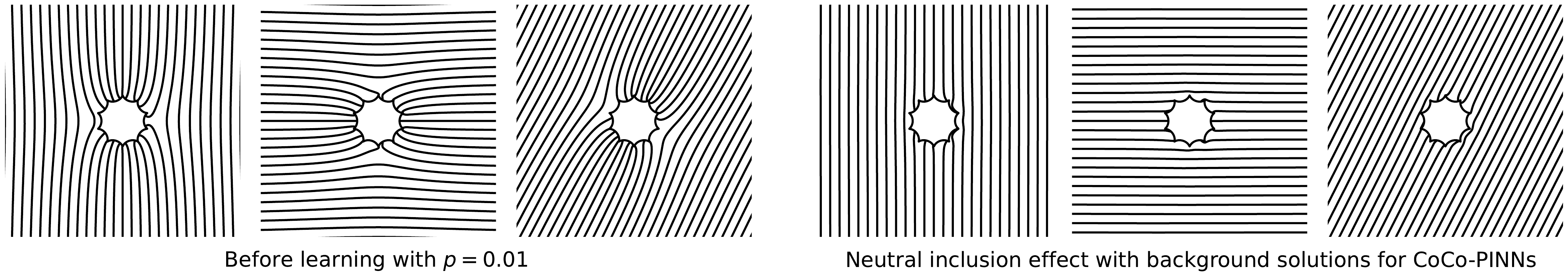}
\caption{
Neutral inclusion effect after training: For the square', fish', kite', spike’ shaped inclusions, the interface function $p^{(n)}$ is simultaneously trained using the two background solutions $H_u(x) = x_1$ and $H_v(x) = x_2$ through the CoCo-PINNs framework.
}
\label{fig:neutral:various}
\end{figure}
%
%

\subsection{Credibility of classical PINNs and CoCo-PINNs}\label{sec:cred}

\begin{table}
\centering
\caption{Credibility results.}
\label{Table2}
\begin{tabular}{cccc}
\hline
&\multirow{2}{*}{Shape}& \multicolumn{2}{c}{Credibility}\\
\cline{3-4}
&&$\|u^{\text{ext}}_{\text{NN}}-u_p\|_{\text{Cred}}$&$\|u^{\text{ext}}_{\text{NN}}-u_p\|_{\infty}$\\
\hline   
\multirow{4}{*}{\shortstack{Classical\\PINNs}}
&square& 2.843e-03 & 1.371e-01\\
&fish& 4.432e-03 & 1.546e-01\\
&kite& 1.246e-03 & 2.483e-01\\
&spike& 8.798e-03 & 4.301e-01\\
\hline
\multirow{4}{*}{\shortstack{CoCo-\\PINNs}}
&square&2.802e-03 & \textbf{9.244e-02}\\
&fish& \textbf{7.468e-04} & \textbf{1.067e-01}\\
&kite& \textbf{4.299e-04} & 2.280e-01\\
&spike& 9.320e-03 & 4.038e-01\\
\hline
\end{tabular}
\end{table}

We investigate the credibility of the two methods. We examine whether the exterior part of the trained forward solution $u^{\text{ext}}_{\text{NN}}$ matches the analytic solution $u_p$.
Recall that we denote $u_p$ as the analytic solution when the coefficients of the interface function $p$ are given by training.
Once training is complete, CoCo-PINNs provide the expansion coefficients of the interface function directly.
For classical PINNs, where the interface function is represented by neural networks, we compute the Fourier coefficients of $p_{\text{NN}}$.
We use the Fourier series expansion up to a sufficiently high order to ensure that the difference between the neural network-designed interface function and its Fourier series is small (see \Cref{fig:Fourier} in \Cref{sec:Fourier:fittings}).
We conduct experiments with four inclusion shapes, as shown in \Cref{fig:shapes}, and the corresponding results are presented subsequently in \Cref{Fig:App:Ex:cre:classic,Fig:App:Ex:cre:CoCo} (\Cref{Sec:App:Ex:cre}).

\Cref{Table2} demonstrates that CoCo-PINNs exhibit meaningfully better performance in the shapes of `fish', `kite', and `spike' compared to classical PINNs.
Although the credibility error for classical PINNs appears similar to that for CoCo-PINNs on the `square' shape as shown in \Cref{Table2}, the trained forward solution by classical PINNs illustrates an exorbitant large deviation that does not coincide with the analytic solution derived from the inverse parameter result, as shown in \Cref{fig:cre:ex}. 
This indicates that, despite its strong performance in minimizing the loss function, the classical PINNs approach fails to effectively function as a forward solver.

\begin{figure}[t!]
\centering
\begin{subfigure}[b]{0.45\textwidth}
\includegraphics[width=\textwidth]{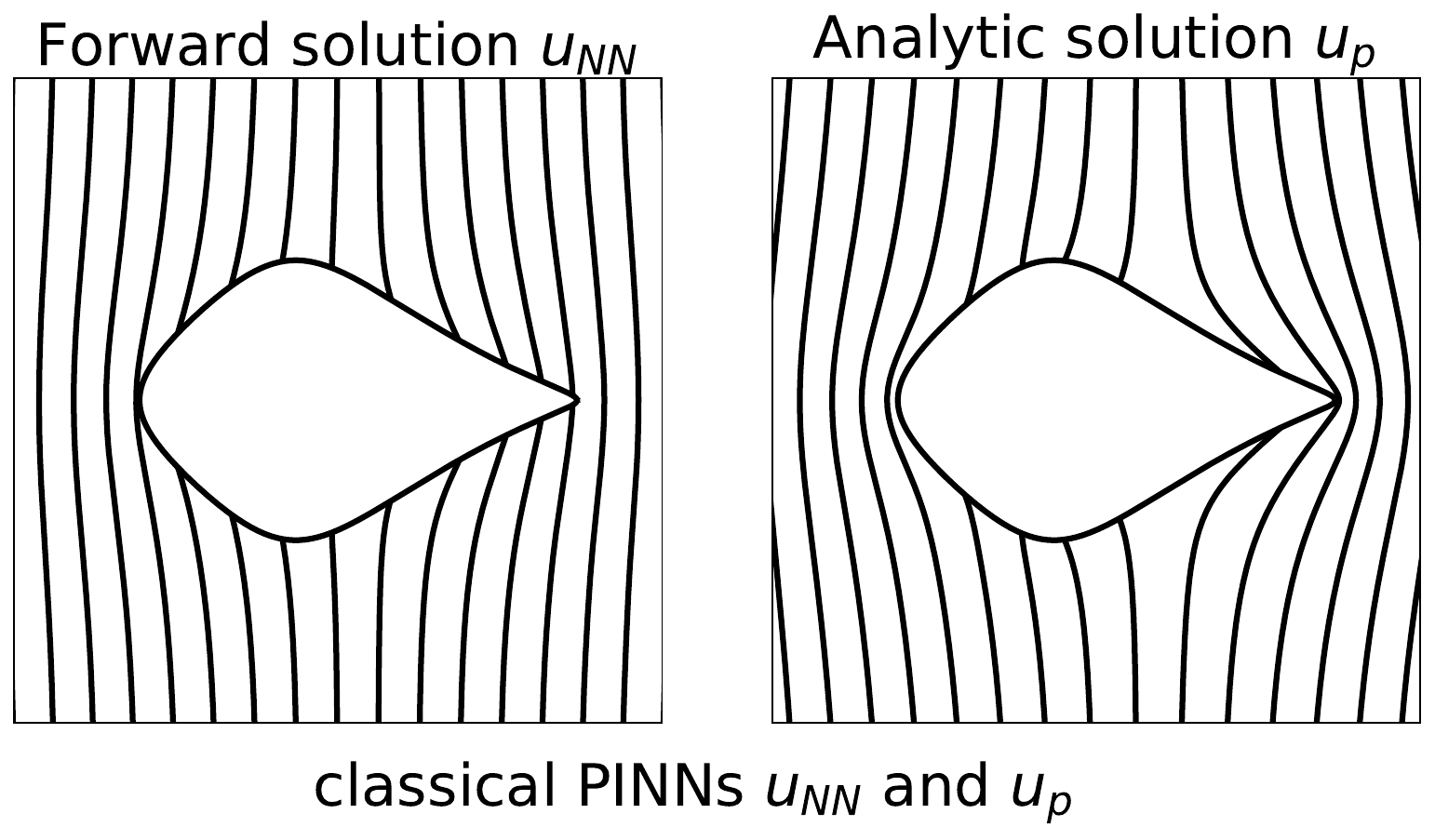}
\end{subfigure}
\hfil
\begin{subfigure}[b]{0.45\textwidth}
\includegraphics[width=\textwidth]{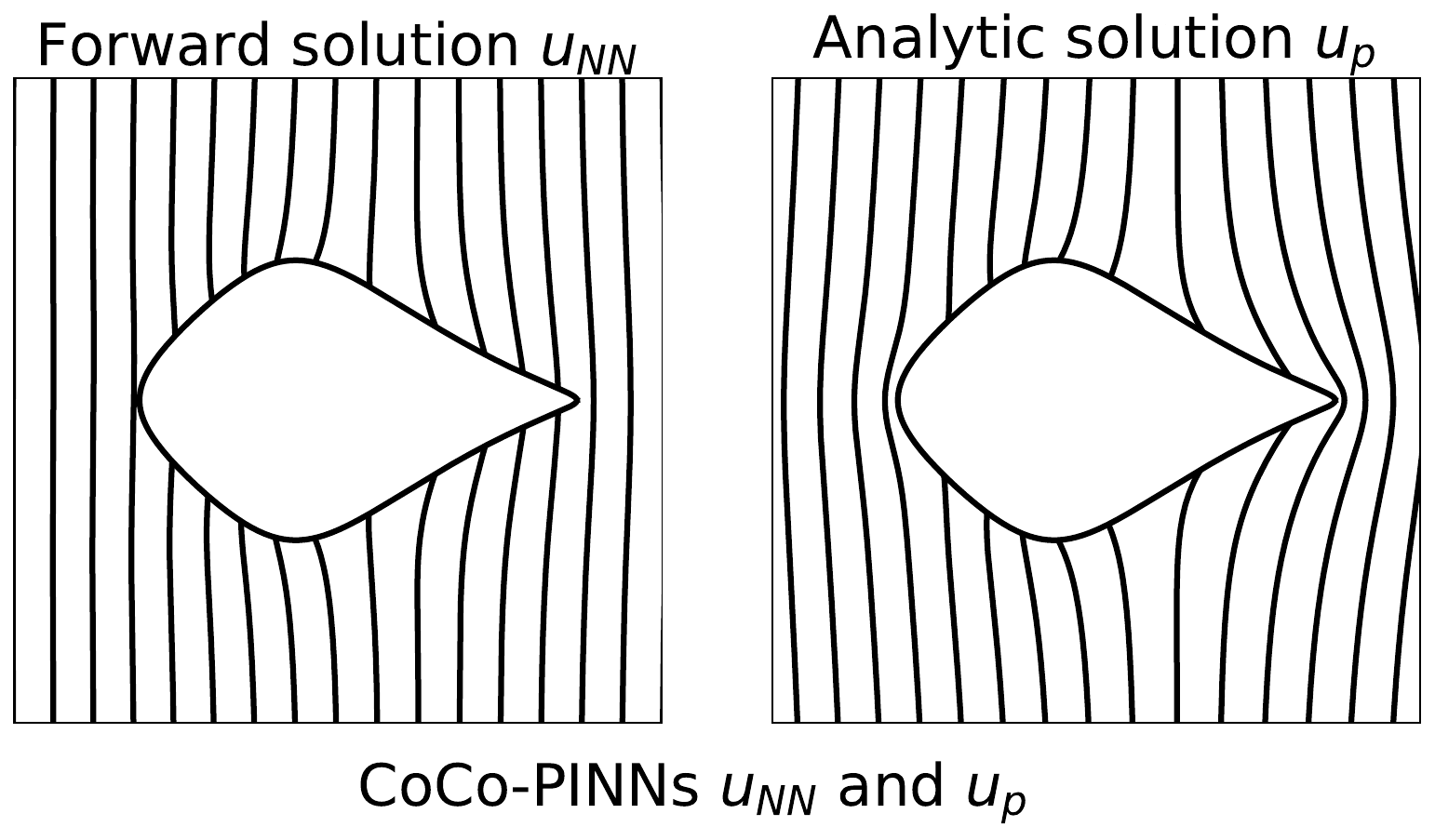}
\end{subfigure}
\caption{
Comparison of credibility: For the fish-shaped inclusion, the trained forward solution $u_{\text{NN}}$ obtained by classical PINNs and CoCo-PINNs is compared with the analytic solution $u_p$. The closer $u_{\text{NN}}$ approximates $u_p$, the more credible the method is considered to be.
}
\label{fig:cre:ex}
\end{figure}

\subsection{Consistency of classical PINNs and CoCo-PINNs}\label{sec:repro}
In this subsection, we examine whether repeated experiments consistently yield similar results.
\Cref{fig:repro} shows the interface functions after training the classical PINNs and CoCo-PINNs, performed independently multiple times. We repetitively test $30$ times under the same condition and plot the interface function pointwise along the boundary of the unit disk.
The blue-dashed and red-bold lines represent the mean of the interface functions produced by classical PINNs and CoCo-PINNs, respectively, while the blue- and red-shaded regions indicate the pointwise standard deviations of the interface functions, respectively.
Each column corresponds to an experiment with `square', `fish', `kite', and `spike'.

\begin{figure}[htbp!]
\centering
\includegraphics[width=\textwidth]{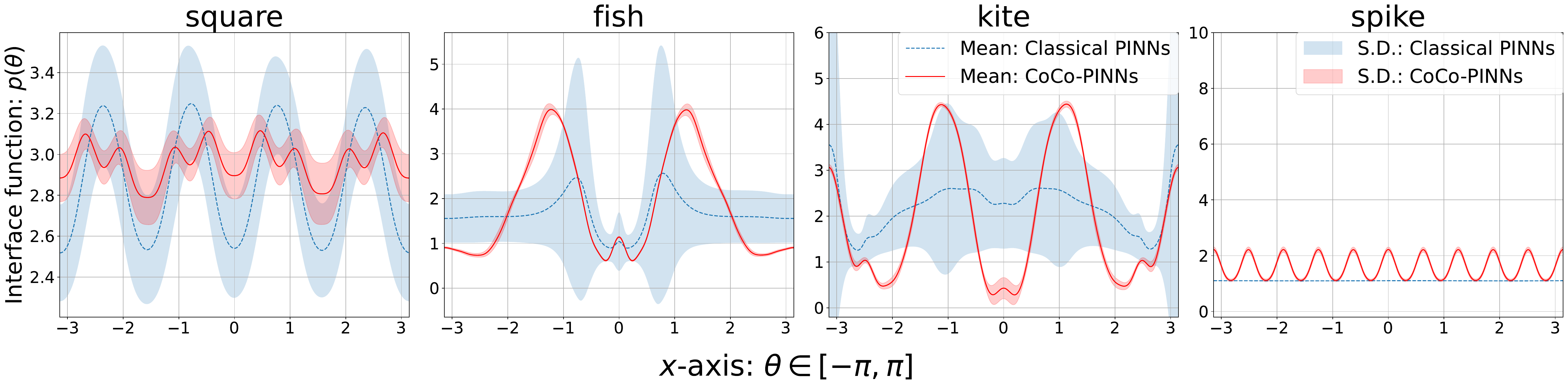}
\caption{
Consistency of interface functions: Repeated reconstructions of the interface function obtained by CoCo-PINNs and classical PINNs. The results demonstrate that CoCo-PINNs produce more consistent interface functions across trials compared to classical PINNs.
}
\label{fig:repro}
\end{figure}

As shown in \Cref{fig:repro}, the interface functions trained with classical PINNs are consistent only for the spike-shaped example, whereas CoCo-PINNs achieve consistency across all examples.
The precise value for the mean of the standard deviations is provided in column $3$ of \Cref{Table3}.
Additionally,  columns $4$ through $6$ of \Cref{Table3} present the three quantities in \Cref{loss:credi,loss:credi2,loss:credi3}, which show the credibility and the errors indicating the neutral inclusion effect.
These results clearly demonstrate that CoCo-PINNs exhibit superior credibility compared to classical PINNs.
Furthermore, CoCo-PINNs outperform classical PINNs in achieving the neutral inclusion effect, particularly for complex shapes, as verified by the quantity $\|u_p-H\|_{\text{P-Neutral}}$. 
In some cases, the trained forward solution $u_{\text{NN}}$ and the inverse solution $p_{\text{NN}}$ produced by classical PINNs show a significant discrepancy, with $u_{\text{NN}}$ differing substantially from the analytic solution derived from $p_{\text{NN}}$ (see \Cref{Fig:App:Ex:cre:classic} in \Cref{Sec:App:Ex:cre}).

In \Cref{fig:repro}, classical PINNs again yield nearly constant-valued solutions --just as they do for the disk-- and therefore fail to effectively capture the problem for the shape with periodic spikes.

\begin{table}[htbp!]
\centering
\caption{Mean of the standard deviation of the interface function, and mean and standard deviations of errors for fitting the background fields after training.
CoCo-PINNs show superior results to classical PINNs for complex shapes such as `fish', `kite', and `spike'. The `Mean of S.D.' denotes the mean of the standard deviation for each point, and `S.D.' denotes the standard deviation.}
 \label{Table3}
\resizebox{\textwidth}{!}{
\begin{tabular}{ccccccccc}
\hline
&\multirow{2}{*}{Shape}&{Interface function}&\multicolumn{2}{c}{$\|u^{\text{ext}}_{\text{NN}}-u_p\|_{\text{Cred}}$}&\multicolumn{2}{c}{$\|u^{\text{ext}}_{\text{NN}}-u_p\|_{\infty}$}&\multicolumn{2}{c}{$\|u_p-H_u\|_{\text{P-Neutral}}$}\\
&&Mean of S.D.&Mean&S.D.&Mean&S.D.&Mean&S.D.\\\hline
\multirow{4}{*}{\shortstack{Classical\\PINNs}}
&square& 2.828e-01 & 2.645e-03 & 4.383e-04 & 1.286e-01 & 2.013e-02 & 4.458e-03 & 6.529e-04\\
&fish & 9.217e-01 & 2.358e-03 & 2.787e-03 & 1.366e-01 & 3.835e-02 & 3.635e-03 & 4.211e-03\\
&kite & 1.199e+00 & 6.607e-02 & 3.327e-01 & 1.356e+00 & 3.573e+00 & 6.872e-02 & 3.485e-01\\
&spike& 9.038e-03 & 8.429e-03 & \textbf{8.331e-05} & 4.129e-01 & \textbf{3.989e-03} & 3.076e-04 & \textbf{2.372e-05}\\
\hline
\multirow{4}{*}{\shortstack{CoCo-\\PINNs}}
&square& \textbf{1.013e-01} & 2.868e-03 & \textbf{9.634e-05} & \textbf{9.516e-02} & \textbf{2.278e-03} & 4.858e-03 & \textbf{1.682e-04}\\
&fish & \textbf{6.646e-02} & \textbf{7.475e-04} & \textbf{3.037e-05} & 1.080e-01 & \textbf{2.383e-03} & \textbf{1.093e-03} & \textbf{4.563e-05}\\
&kite & \textbf{1.009e-01} & \textbf{4.181e-04} & \textbf{8.149e-05} & \textbf{2.180e-01} & \textbf{2.505e-02} &\textbf{ 6.031e-04} & \textbf{1.356e-04}\\
&spike& \textbf{7.068e-02} & 9.518e-03 & 3.681e-04 & 4.077e-01 & 6.375e-03 & 3.441e-04 & 1.440e-04\\
\hline          
\end{tabular}
}
\end{table}

\subsection{Stability of classical PINNs and CoCo-PINNs}
In this subsection, we assess the stability of the interface function along with the change of environments of PDEs.
Since both classical PINNs and CoCo-PINNs are trained for a fixed domain $\Omega$ and background field $H$, we focus on stability with respect to different conductivities $\sigma_c$.
In \Cref{fig:stability}, we present the experimental results for various conductivity contrasts, where $\sigma_c = 3, 4, 5, 6, 7$ and $\sigma_m = 1$. 
The experiments were conducted for four different inclusion shapes: `square', `fish', `kite', and `spike'.
Recall that the ill-posed nature of inverse problems can lead to significant instability, causing the inverse solution to exhibit large deviations in response to environmental changes or re-experimentation. 
Classical PINNs for neutral inclusions with imperfect conditions represent such an unstable case, as demonstrated in \Cref{fig:stability}.
In contrast, our CoCo-PINNs are stable for repeated experiments, and we confirmed that CoCo-PINNs are stable for slightly changed environments.
\Cref{Table4} provides the mean of standard deviations for all experiments used in \Cref{fig:stability}.
CoCo-PINNs demonstrate substantially improved consistency and stability over classical PINNs, as evidenced in \Cref{Table4}.

\begin{figure}[t!]
\centering
\includegraphics[width=\textwidth]{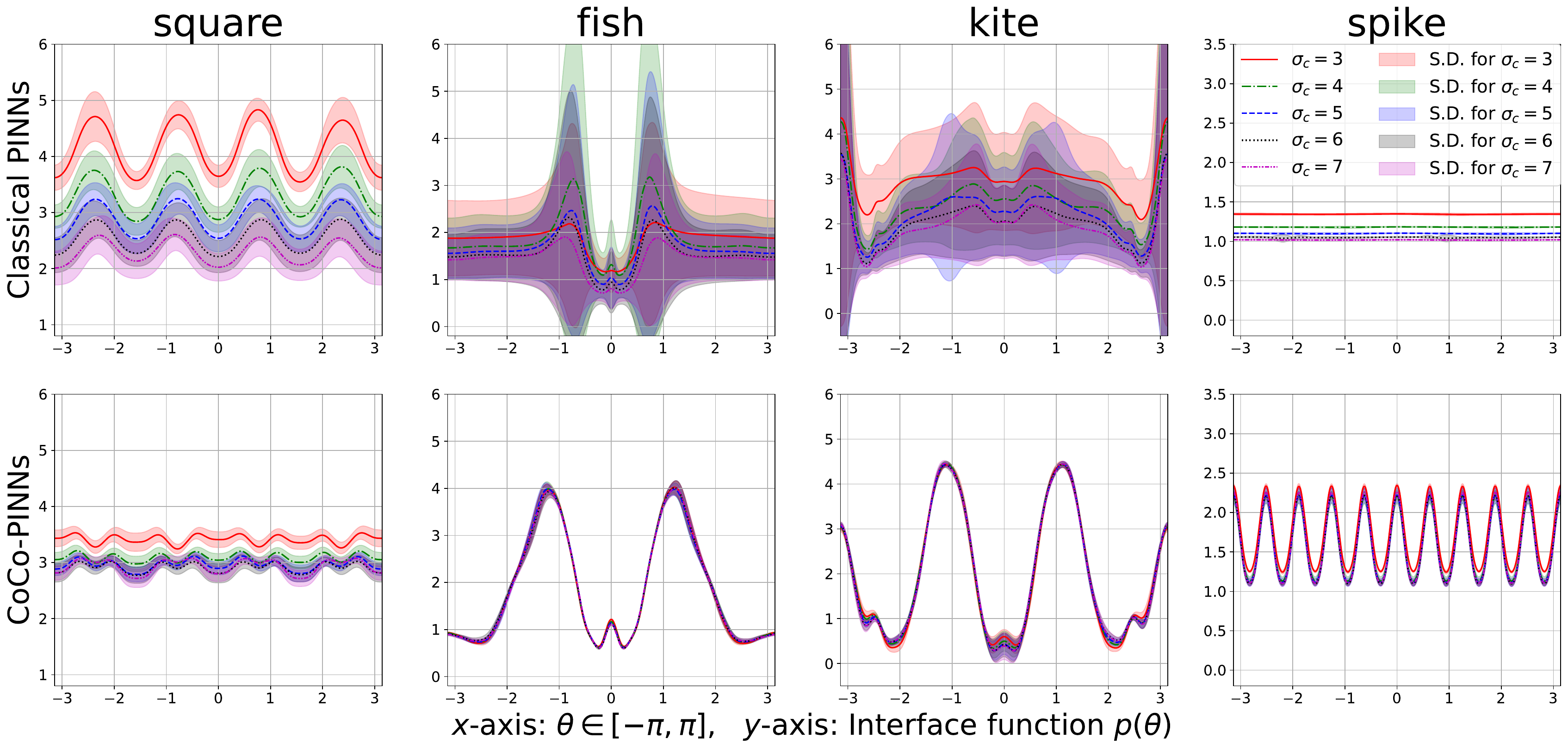}
\caption{
Stability of interface functions under varying conductivities:
The first and second rows display the interface functions obtained from classical PINNs and CoCo-PINNs, respectively. 
The mean values of the standard deviations are reported in \Cref{Table4}. 
Across all cases, the interface functions learned by CoCo-PINNs exhibit significantly greater stability with respect to variations in conductivity compared to those from classical PINNs.
}
\label{fig:stability}
\end{figure}

\begin{table}[htbp!]
\centering
\caption{Mean of standard deviations of interface functions for stability experiments.}\label{Table4}
\begin{tabular}{ccccccc}
\hline
&\multirow{2}{*}{Shape}&\multicolumn{5}{c}{Conductivities for interior}\\
\cline{3-7}
&&$\sigma_c=3$&$\sigma_c=4$&$\sigma_c=5$&$\sigma_c=6$&$\sigma_c=7$\\\hline
\multirow{4}{*}{\shortstack{Classical\\PINNs}}
&square& 2.679e-01 & 2.944e-01 & 2.828e-01 & 3.410e-01 & 3.316e-01\\
&fish & 9.699e-01 & 1.208e+00 & 9.217e-01 & 9.211e-01 & 7.013e-01\\
&kite & 1.278e+00 & 1.131e+00 & 1.199e+00 & 9.786e-01 & 9.539e-01\\
&spike& \textbf{1.281e-02} & \textbf{8.433e-03} & \textbf{9.038e-03} & \textbf{1.364e-02} & \textbf{7.030e-03}\\
\hline     
\multirow{4}{*}{\shortstack{CoCo-\\PINNs}}
&square& \textbf{1.200e-01} & \textbf{1.057e-01} & \textbf{1.013e-01} & \textbf{1.240e-01} & \textbf{1.176e-01} \\
&fish & \textbf{6.505e-02} & \textbf{7.071e-02} & \textbf{6.646e-02} & \textbf{7.784e-02} & \textbf{7.521e-02}\\
&kite & \textbf{1.139e-01} & \textbf{1.016e-01} & \textbf{1.009e-01} & \textbf{1.135e-01} & \textbf{1.153e-01}\\
&spike& 2.379e-02 & 7.489e-02 & 7.068e-02 & 4.121e-02 & 5.800e-02\\
\hline
\end{tabular}
\end{table}

\section{Further experiments}\label{app:sec:NN}
This section presents additional experimental results to assess the neutral inclusion effect, demonstrate the credibility for the four shapes defined in \Cref{fig:shapes}, and compare the training time of both methods.

\subsection{Neutral inclusion effect for arbitrary fields}\label{app:Neutral_exs}
In this subsection, we present the neutral inclusion effect of shapes and different fields $H(x_1, x_2) = x_1, x_2$, and $2x_1-x_2$.
The neutral inclusion effects for one random experiment result are given in \Cref{Table5}.

\begin{table}[htbp!]
\centering
\caption{Errors for fitting the background fields after neutral inclusion}
\label{Table5}
\renewcommand{\arraystretch}{1.2}
\begin{tabular}{ccc c c}
\hline
&\multirow{2}{*}{Shape}& \multicolumn{3}{c}{$\|u_p-H\|_{\text{P-Neutral}}$ with} \\
\cline{3-5} 
&&$H(x)=x_1$ & $H(x)=x_2$ & $H(x)=2x_1-x_2$ \\
\hline
\multirow{4}{*}{\shortstack{CoCo-\\PINNs}}
&square& 4.738e-03 & 4.981e-03 & 2.395e-02\\
&fish& 1.113e-03 & 2.568e-04 & 4.491e-03\\
&kite& 4.664e-04 & 3.839e-04 & 2.361e-03\\
&spike& 3.823e-04 & 3.858e-04 & 1.937e-03\\
\hline
\end{tabular}
\end{table}

After many fair experiments with both CoCo-PINNs and classical PINNs, we concluded that CoCo-PINNs were superior.
After that, we tested the neutral inclusion effect for each shape by utilizing the CoCo-PINNs.
\Cref{fig:neutral:various} presents the results.
In the case of unit circle inclusion, exact neutral inclusion appeared.
The shape we used may have no interface functions that make exact neutral inclusions.
Notwithstanding, the CoCo-PINNs results for the neutral inclusion effect are, to some extent, satisfactory.

\subsection{Illustrations for credibility}\label{Sec:App:Ex:cre}
All experiments described in \Cref{Table2} are illustrated in \Cref{Fig:App:Ex:cre:CoCo,Fig:App:Ex:cre:classic}, by utilizing CoCo-PINNs and classical PINNs, respectively.
We illustrate the pairs of 
$$
\left(u_{\text{NN}}, u_p, \frac{|u_{\text{NN}}^{\text{ext}}-u_p|^2}{|\Omega^{\text{ext}}|}, \frac{|u_{\text{NN}}^{\text{ext}}-H|^2}{|\Omega^{\text{ext}}|}\right)
$$
for each shape in \Cref{fig:shapes}.

\begin{figure}[t!]
\centering
\begin{subfigure}{.7\linewidth}
    \centering
    \includegraphics[width=\linewidth]{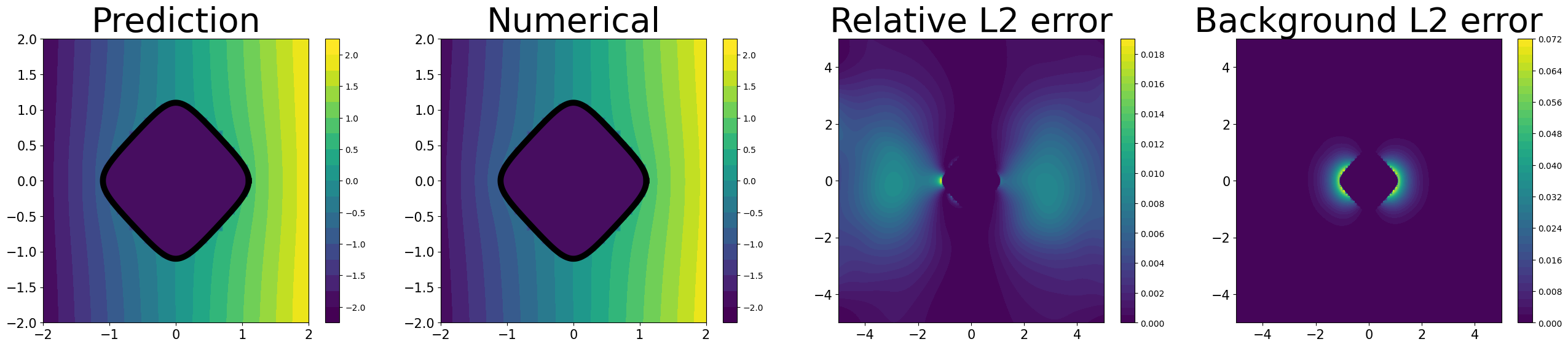}
\end{subfigure}
\\
\begin{subfigure}{.7\linewidth}
    \centering
    \includegraphics[width=\linewidth]{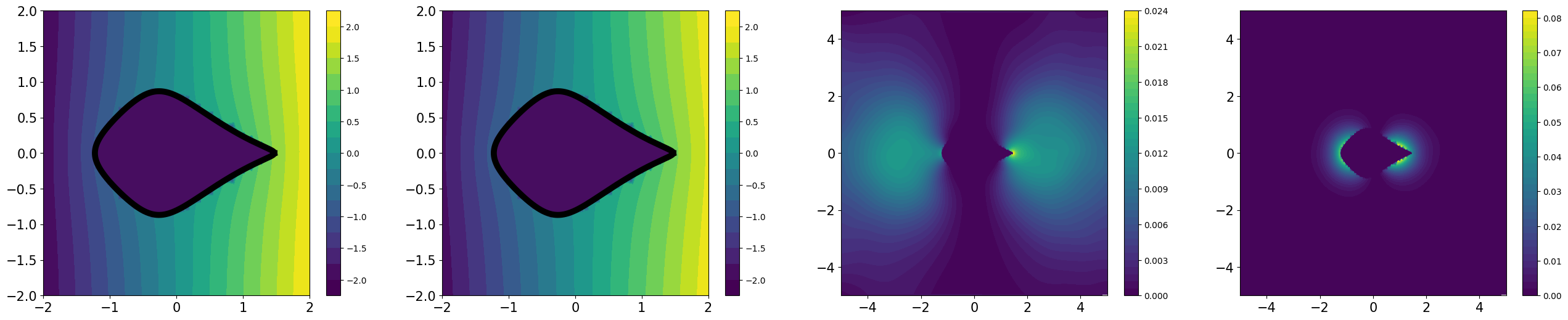}
\end{subfigure}
\\
\begin{subfigure}{.7\linewidth}
    \centering
    \includegraphics[width=\linewidth]{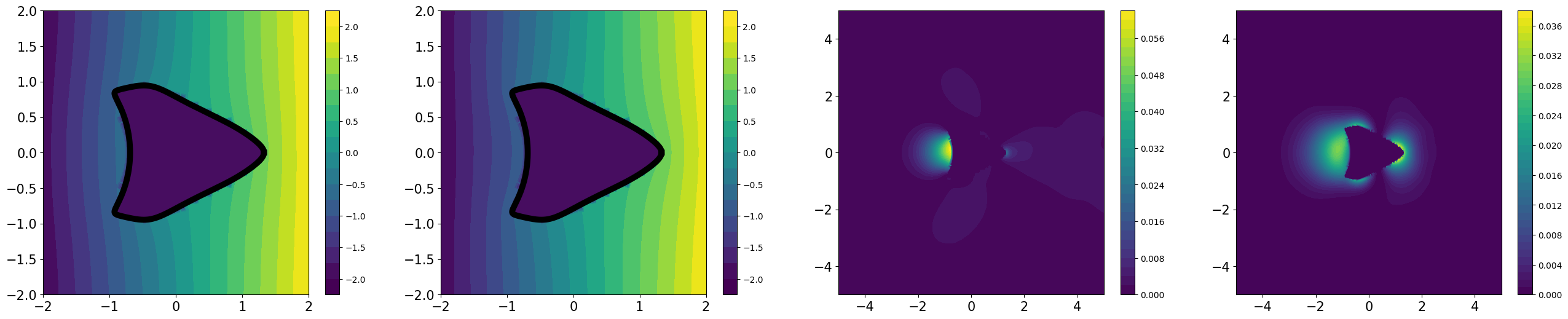}
\end{subfigure}
\\
\begin{subfigure}{.7\linewidth}
    \centering
    \includegraphics[width=\linewidth]{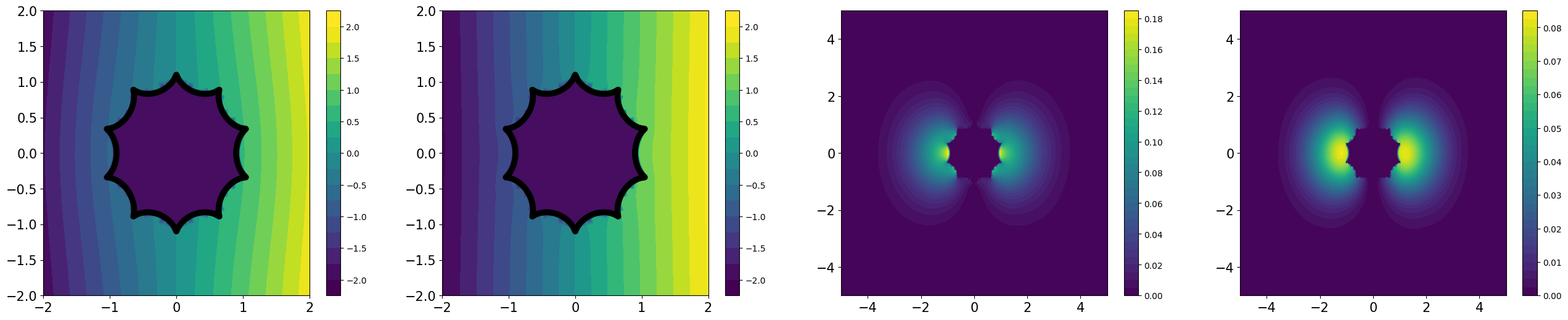}
\end{subfigure}
\caption{
Experiment results using classical PINNs: For kite and spike shapes, notable discrepancies are observed between the forward solutions obtained via classical PINNs and the corresponding analytic solutions.
}
\label{Fig:App:Ex:cre:classic}
\end{figure}

\begin{figure}[htbp!]
\centering
\begin{subfigure}{.7\linewidth}
    \centering
    \includegraphics[width=\linewidth]{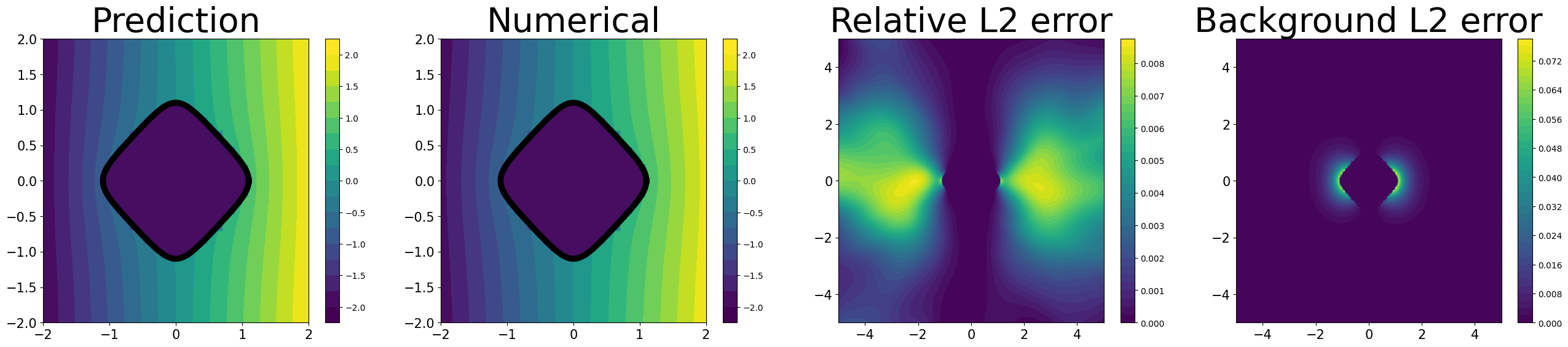}
\end{subfigure}
\\
\begin{subfigure}{.7\linewidth}
    \centering
    \includegraphics[width=\linewidth]{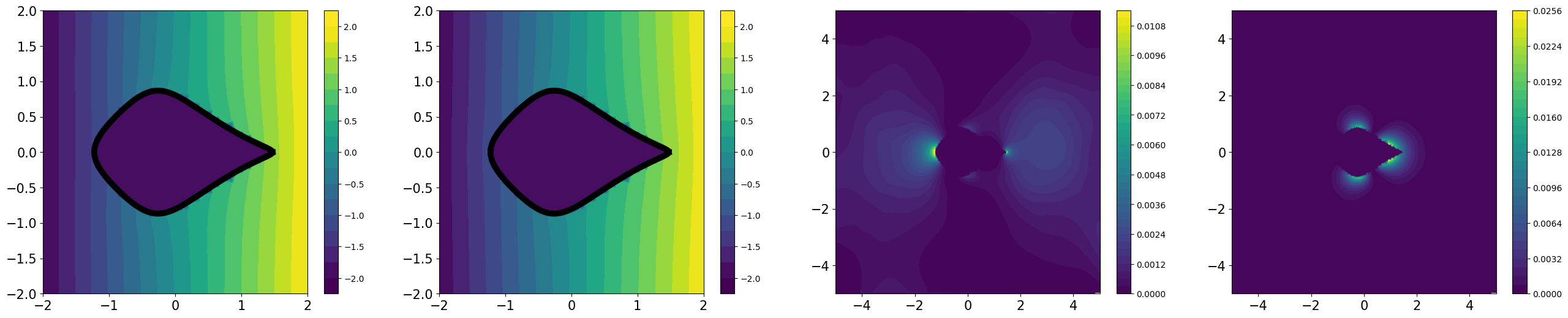}
\end{subfigure}
\\
\begin{subfigure}{.7\linewidth}
    \centering
    \includegraphics[width=\linewidth]{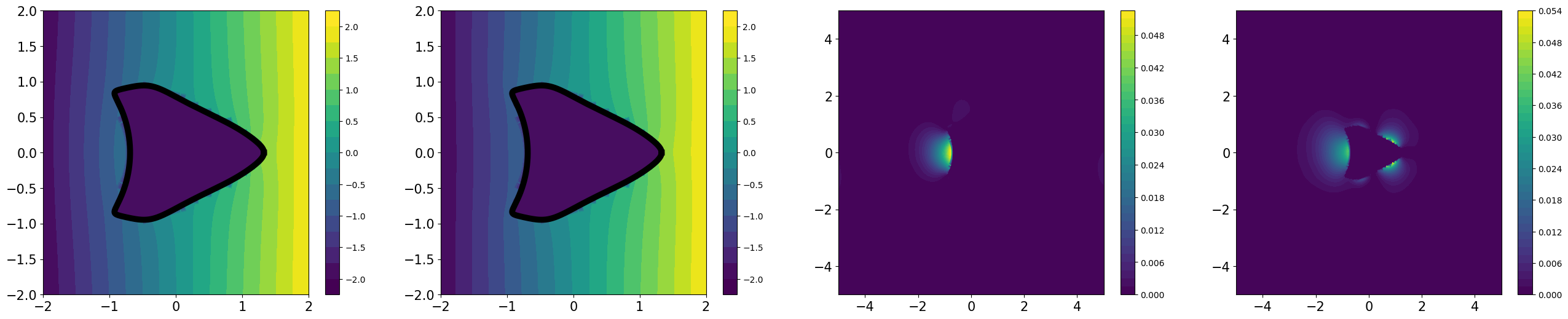}
\end{subfigure}
\\
\begin{subfigure}{.7\linewidth}
    \centering
    \includegraphics[width=\linewidth]{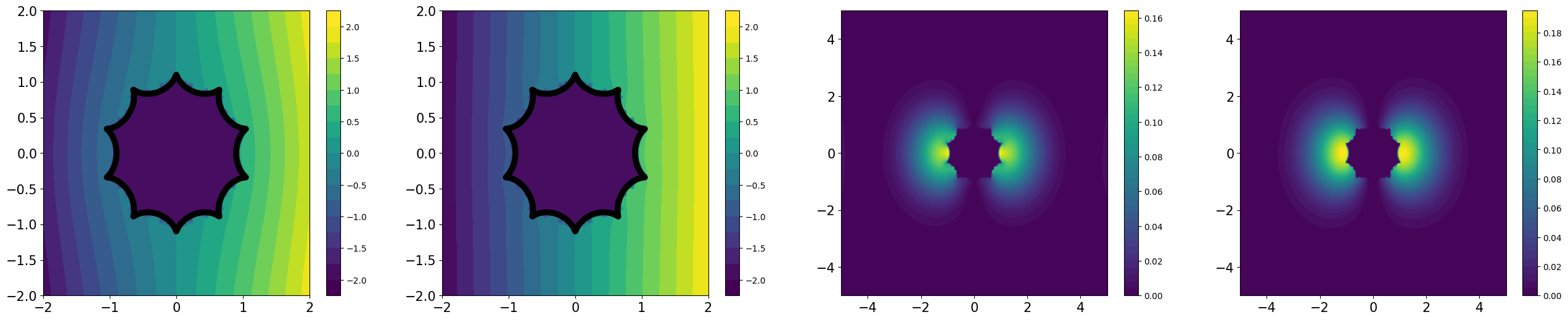}
\end{subfigure}
\caption{
Experiment results using CoCo-PINNs: For all shapes considered, the forward solutions obtained via CoCo-PINNs more closely match the analytic solutions than those obtained by classical PINNs.
}
\label{Fig:App:Ex:cre:CoCo}
\end{figure}

As shown \Cref{Fig:App:Ex:cre:CoCo,Fig:App:Ex:cre:classic}, the trained forward solutions by CoCo-PINNs $u_{\text{NN}}$ are more similar to the analytic solution $u_p$ using the interface parameter given by the CoCo-PINNs' training results than the classical PINNs’ one.

\subsection{Training time}

\begin{table}[t!]\small
\vspace{-0.3cm}
\centering
\caption{Training time.}\label{Table7}
\renewcommand{\arraystretch}{1.2}
\begin{tabular}{ccc}
\hline
&Shape& Time (s): Mean of 30 experiments\\
\hline   
\multirow{4}{*}{\shortstack{Classical\\PINNs}}
&square&  1259.2729 \\
&fish& 1254.7109\\
&kite&  1260.5721\\
&spike & 1257.5664\\
\hline
\multirow{4}{*}{\shortstack{CoCo-\\PINNs}}
&square& 1670.5210 \\
&fish& 1671.7479 \\
&kite& 1667.0469 \\
&spike& 1668.4443 \\
\hline
\end{tabular}
\vspace{-0.3cm}
\end{table}

\Cref{Table7} presents a simple comparison of two methods in terms of computation time. We use a CPU (Intel® Xeon® Gold 6530) and a GPU (NVIDIA RTX 6000 Ada Generation). The learning time for CoCo-PINNs’ results is approximately 1.3 times that of classical PINNs’ results.
The reported training times are averaged over 30 repeated experiments.

\subsection{Ablation experiments on the regularization term for CoCo-PINNs}
We examined the impact of the regularization term in the loss function in \Cref{loss:reg}. 
The ablation experiments demonstrate that regularization generally enhances the consistency of the recovered interface function $p$ for CoCo-PINNs. Specifically, the regularized model exhibits clearly reduced variability compared to the unregularized model (see Figure \ref{fig:comparison_fish_spike}). Also, in experiments for spike geometries, we observed that the regularized model maintains more consistent periodicity and amplitude in the oscillatory interface function (see also Figure \ref{fig:comparison_fish_spike}).

\begin{figure}[t!]
\centering

\begin{subfigure}[t]{0.48\textwidth}
    \centering
    \includegraphics[width=\textwidth]{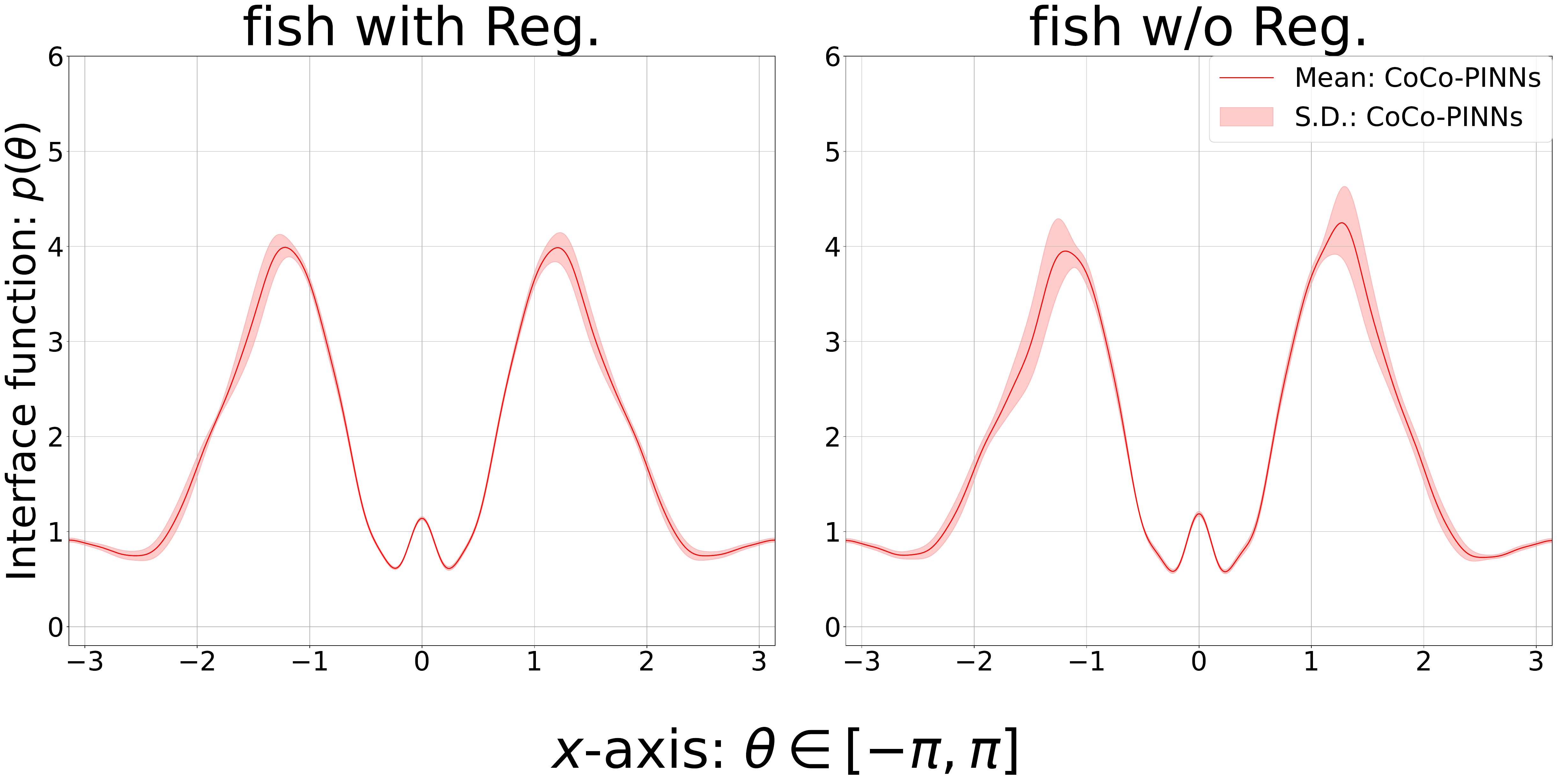}
\end{subfigure}
\hfill
\begin{subfigure}[t]{0.48\textwidth}
    \centering
    \includegraphics[width=\textwidth]{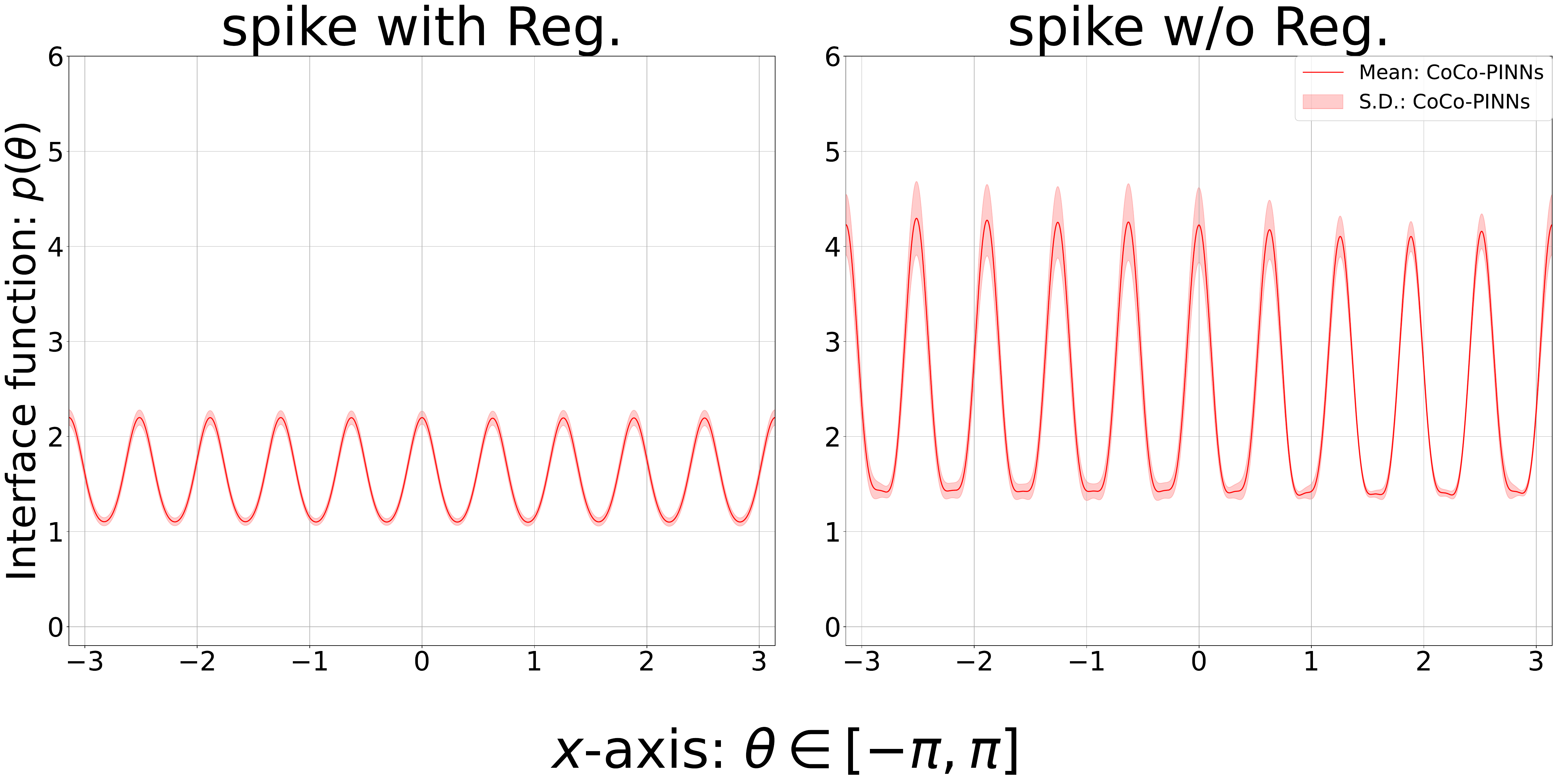}
\end{subfigure}
\caption{
Consistency comparison of CoCo-PINNs with and without regularization for fish- and spike-shaped inclusion.}
\label{fig:comparison_fish_spike}
\end{figure}

\section{Conclusion}
We focus on the inverse problem of identifying an imperfect function that makes a given simply connected inclusion a neutral inclusion. 
We introduce a novel approach of Conformal mapping Coordinates Physics-Informed Neural Networks (CoCo-PINNs) based on complex analysis and PDEs.
Our proposed approach of CoCo-PINNs solves both the forward and inverse problems simultaneously, much more effectively than the classical PINNs approach.
While the classical PINNs approach may occasionally demonstrate success in finding an imperfect function with a strong neutral inclusion effect, the reliability of this performance remains uncertain.
In contrast, CoCo-PINNs present high credibility, consistency, and stability, with the additional advantage of being explainable through analytical results. 
The potential applications of this method extend to analyzing and manipulating the interaction of embedded inhomogeneities and surrounding media, such as finding inclusions having uniform fields in their interiors.
Several questions remain, including the generalization to multiple inclusions and three-dimensional problems, as well as proving the existence of an interface function that achieves neutrality.

The primary motivation for using conformal mapping is to establish a coordinate system that facilitates the automatic generation and classification of collocation points. 
Furthermore, employing these coordinates allows for a direct comparison between our numerical solution and the analytic solution, which is formulated in conformal mapping coordinates. It is worth noting that conformal maps may induce length distortion; for shapes causing severe distortion, an adaptive collocation strategy may be necessary to maintain accuracy. While conformal mappings generally do not exist for arbitrary shapes in three dimensions, the core strategy--establishing a method that naturally defines normal directions and distinct interior/exterior domains--remains applicable. One potential approach is the use of machine learning methods, such as Deep Signed Distance Function (DeepSDF) models \cite{Park:2019:DeepSDF}.

\appendix
\section{Geometric function theory}\label{app:ex_conf}
Geometric function theory is the research area of mathematics with the corresponding geometric properties of analytic functions.
One remarkable result is the Riemann mapping theorem.
We briefly introduce this theorem with related results.
A connected open set in the complex plane is called a domain. 
We say that a domain $\Omega$ is simply connected if its complement $\mathbb{C}\setminus\Omega$ is connected. 
\begin{theorem}[Riemann mapping theorem] \label{thm:Riemann}
If $\Omega\subsetneq\mathbb{C}$ is a nonempty simply connected domain, then there exists a conformal map from the unit ball $B=\{z\in\mathbb{C}:|z|<1\}$ onto $\Omega$.
\end{theorem}

We assume that $\Omega\subsetneq\mathbb{C}$ is a nonempty simply connected bounded domain. Then, by the Riemann mapping theorem, there exists a unique $\gamma>0$ and conformal mapping $\Psi$ from $D=\{w\in\mathbb{C}:|w|>\gamma\}$ onto $\mathbb{C}\setminus \overline{\Omega}$ such that $\Psi(\infty)=\infty$, $\Psi'(\infty)=1$, and
\begin{equation}\label{def:conformal}
\Psi(w) = w + a_0 + \frac{a_1}{w}+\frac{a_2}{w^2}+\cdots.
\end{equation}
The quantity $\gamma$ in \Cref{def:conformal} is called the conformal radius of $\Omega$. 
One can obtain \Cref{def:conformal} by using \Cref{thm:Riemann}, the power series expansion of an analytic function, and its reflection with respect to a circle; 
we refer to for instance \cite[Chapter 1.2]{Pommerenke:1992:BBC} for the derivation.

We further assume, in accordance with the assumptions in \cite{Choi:2024:CIV}, that $\Omega$ has an analytic boundary, that is, $\Psi$ can be conformally extended to ${w\in\mathbb{C}:|w|>\gamma-\epsilon}$ for some $\epsilon>0$.

\section*{Acknowledgement}
This work is supported by the National Research Foundation of Korea(NRF) grants funded by the Korea government(MSIT) (RS-2024-00359109, RS-2025-02303239). This work is also supported by the Institute of Information \& Communications Technology Planning \& Evaluation (IITP) grant funded by the Korea government (MSIT) [RS-2021-II211341, Artificial Intelligence Graduate School Program (Chung-Ang University)].

%



\ifx \bblindex \undefined \def \bblindex #1{} \fi\ifx \bbljournal \undefined
  \def \bbljournal #1{{\em #1}\index{#1@{\em #1}}} \fi\ifx \bblnumber
  \undefined \def \bblnumber #1{{\bf #1}} \fi\ifx \bblvolume \undefined \def
  \bblvolume #1{{\bf #1}} \fi\ifx \noopsort \undefined \def \noopsort #1{} \fi

\end{document}